\documentclass[journal]{IEEEtran}
\usepackage{booktabs}
\usepackage{subfigure}
\usepackage{graphicx}
\usepackage{float}
\usepackage{verbatim}
\usepackage{graphicx,amssymb,mathrsfs,amsmath,latexsym}
\usepackage{epsfig}
\usepackage{epstopdf}
\usepackage{color,xcolor}
\usepackage{algpseudocode}
\usepackage[linesnumbered,ruled]{algorithm2e}
\usepackage{cite}
\usepackage[colorlinks,linkcolor=black,anchorcolor=black,citecolor=black]{hyperref}
\usepackage{multirow}
\usepackage{array}

\newcolumntype{C}[1]{>{\PreserveBackslash\centering}p{#1}}
\newcolumntype{R}[1]{>{\PreserveBackslash\raggedleft}p{#1}}
\newcolumntype{L}[1]{>{\PreserveBackslash\raggedright}p{#1}}

\makeatletter
\def\hlinew#1{%
  \noalign{\ifnum0=`}\fi\hrule \@height #1 \futurelet
   \reserved@a\@xhline}

\newcommand{\PreserveBackslash}[1]{\let\temp=\\#1\let\\=\temp}

\begin{document}
\title{ORSIm Detector: A Novel Object Detection Framework in Optical Remote Sensing Imagery Using Spatial-Frequency Channel Features}

\author{Xin~Wu,
        Danfeng Hong,~\IEEEmembership{Student Member,~IEEE,}
        Jiaojiao Tian,
        Jocelyn Chanussot,~\IEEEmembership{Fellow,~IEEE,}
        Wei Li,~\IEEEmembership{Senior Member,~IEEE,}
        Ran Tao,~\IEEEmembership{Senior Member,~IEEE}
\thanks{This work was supported, in part by the National Natural Science Foundation of China under Grant 61331021, and Grant 61421001, in part by the National Natural Science Foundation of China (U1833203). (\emph{Corresponding author: Danfeng Hong.}) }
\thanks{X. Wu, W. Li and R. Tao are with the School of Information and Electronics, Beijing Institute of Technology 100081, China, and  100081, ChinBeijing Key Laboratory of Fractional Signals and Systems, School of Information and Electronics, Beijing Institute of Technology, Beijinga.(e-mail: aixueshuqian@gmail.com, liwei089@ieee.org, rantao@bit.edu.cn).}%
\thanks{D. Hong is with the Remote Sensing Technology Institute (IMF), German Aerospace Center (DLR), 82234 Wessling, Germany, and Signal Processing in Earth Observation (SiPEO), Technical University of Munich (TUM), 80333 Munich, Germany. (e-mail: danfeng.hong@dlr.de)}
\thanks{J. Tian is with the Remote Sensing Technology Institute (IMF), German Aerospace Center (DLR), 82234 Wessling, Germany. (e-mail: jiaojiao.tian@dlr.de)}
\thanks{J. Chanussot is with the Univ. Grenoble Alpes, CNRS, Grenoble INP, GIPSA-lab, F-38000 Grenoble, France, also with the Faculty of Electrical and Computer Engineering, University of Iceland, Reykjavik 101, Iceland. (e-mail:  jocelyn@hi.is)}
}

\markboth{IEEE Transactions on Geoscience and Remote Sensing,~Vol.~XX, No.~XX,~2019}%
{Shell \MakeLowercase{\textit{et al.}}: Bare Demo of IEEEtran.cls for IEEE Journals}

\maketitle

\begin{abstract}
\textcolor{blue}{This is the pre-acceptance version, to read the final version please go to IEEE Transactions on Geoscience andRemote Sensing on IEEE Xplore.} With the rapid development of spaceborne imaging techniques, object detection in optical remote sensing imagery has drawn much attention in recent decades. While many advanced works have been developed with powerful learning algorithms, the incomplete feature representation still cannot meet the demand for effectively and efficiently handling image deformations, particularly objective scaling and rotation. To this end, we propose a novel object detection framework, called \textbf{o}ptical \textbf{r}emote \textbf{s}ensing \textbf{im}agery detector (ORSIm detector), integrating diverse channel features extraction, feature learning, fast image pyramid matching, and boosting strategy. ORSIm detector adopts a novel spatial-frequency channel feature (SFCF) by jointly considering the rotation-invariant channel features constructed in frequency domain and the original spatial channel features (e.g., color channel, gradient magnitude). Subsequently, we refine SFCF using learning-based strategy in order to obtain the high-level or semantically meaningful features. In the test phase, we achieve a fast and coarsely-scaled channel computation by mathematically estimating a scaling factor in the image domain. Extensive experimental results conducted on the two different airborne datasets are performed to demonstrate the superiority and effectiveness in comparison with previous state-of-the-art methods.
\end{abstract}
%
\IEEEpeerreviewmaketitle

\section{Introduction}
\begin{figure}[!t]
\centering\includegraphics[width=0.8\linewidth]{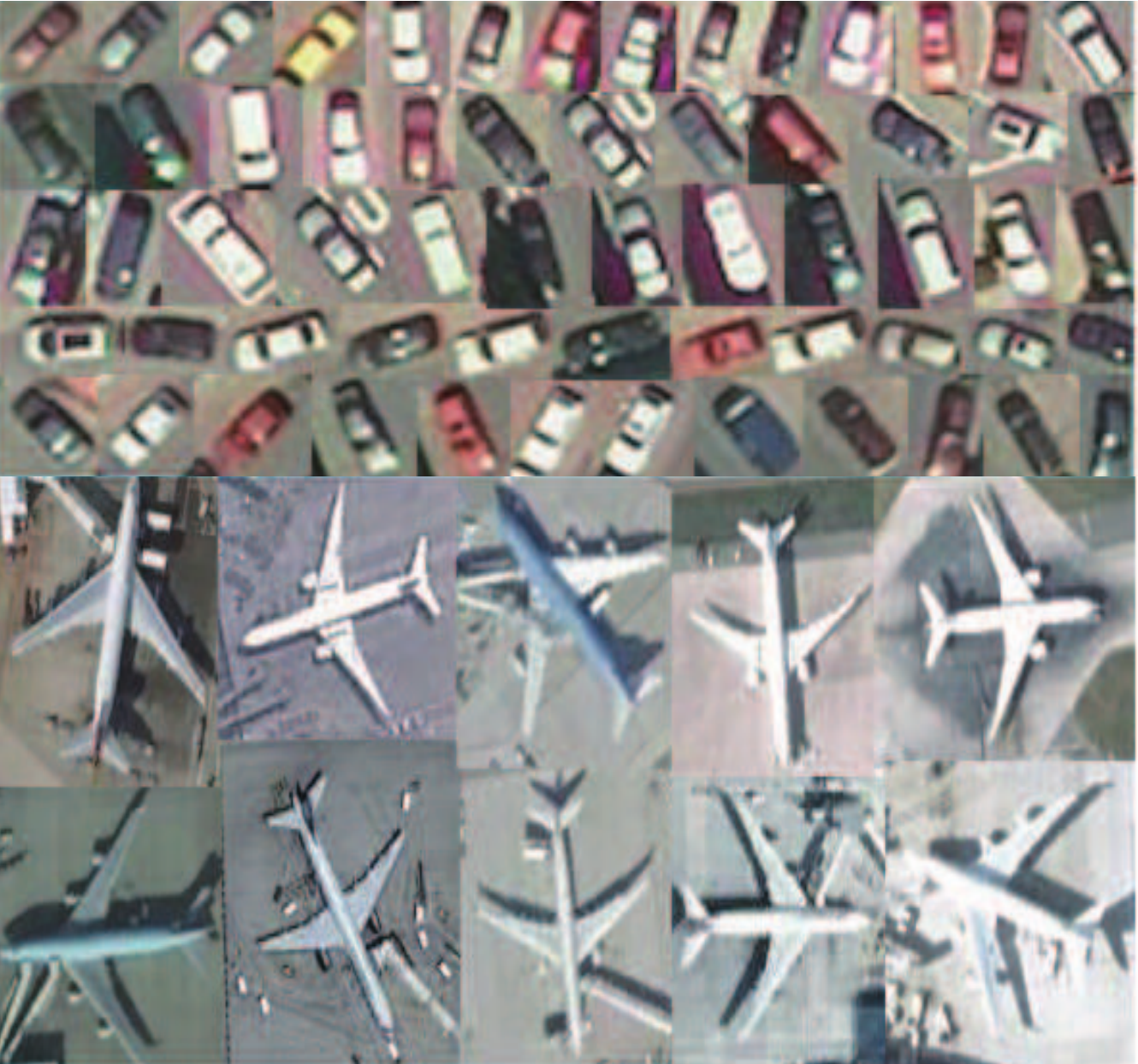}
\caption{Some seeds used in our work for cars and airplanes object detection.}
\label{fig1}
\end{figure}

\IEEEPARstart{G}{enerally} speaking, optical remote sensing imagery is collected from airborne or satellite sources in the range of $400 \sim 760$ nm. As a large amount of multispectral images or very-high-resolution RGB images are freely available on a large scale, there is a growing interest in various applications, such as dimensionality reduction \cite{hong2017learning, hong2018joint}, segmentation \cite{hu2016bilevel,hu2017unsupervised}, unmixing \cite{henrot2016dynamical, hong2017ICIP, hong2018sulora, hong2019augmented}, data fusion \cite{veganzones2016hyperspectral, hong2018cospace, hong2019learnable}, object detection and tracking \cite{yokoya2015object, cheng2016survey, tochon2017object}, and classification or recognition \cite{hong2014improved,li2018regionCNN,li2018CNN,hong2016robust}. In recent years, geospatial object detection has been paid much attention due to its importance in environmental monitoring, ecological protection, hazard responses, etc. However, optical remote sensing imagery inevitably suffers from all kinds of deformations, e.g. variabilities in viewpoint, scaling and direction, which results in performance degradation of detection algorithm. In addition, objects in optical remote sensing imagery \cite{maussang2007higher,ISPRS2008,GRS2014,Corbane2010,GRS2010}, such as cars and airplanes in Fig. \ref{fig1}, are generally small relative to the Ground Sampling Distance (GSD) with cluttered backgrounds. To overcome these challenges, object detection in remote sensing community has been extensively studied since the 1980s.
\begin{figure*}[!t]
\centering\includegraphics[width=0.95\linewidth]{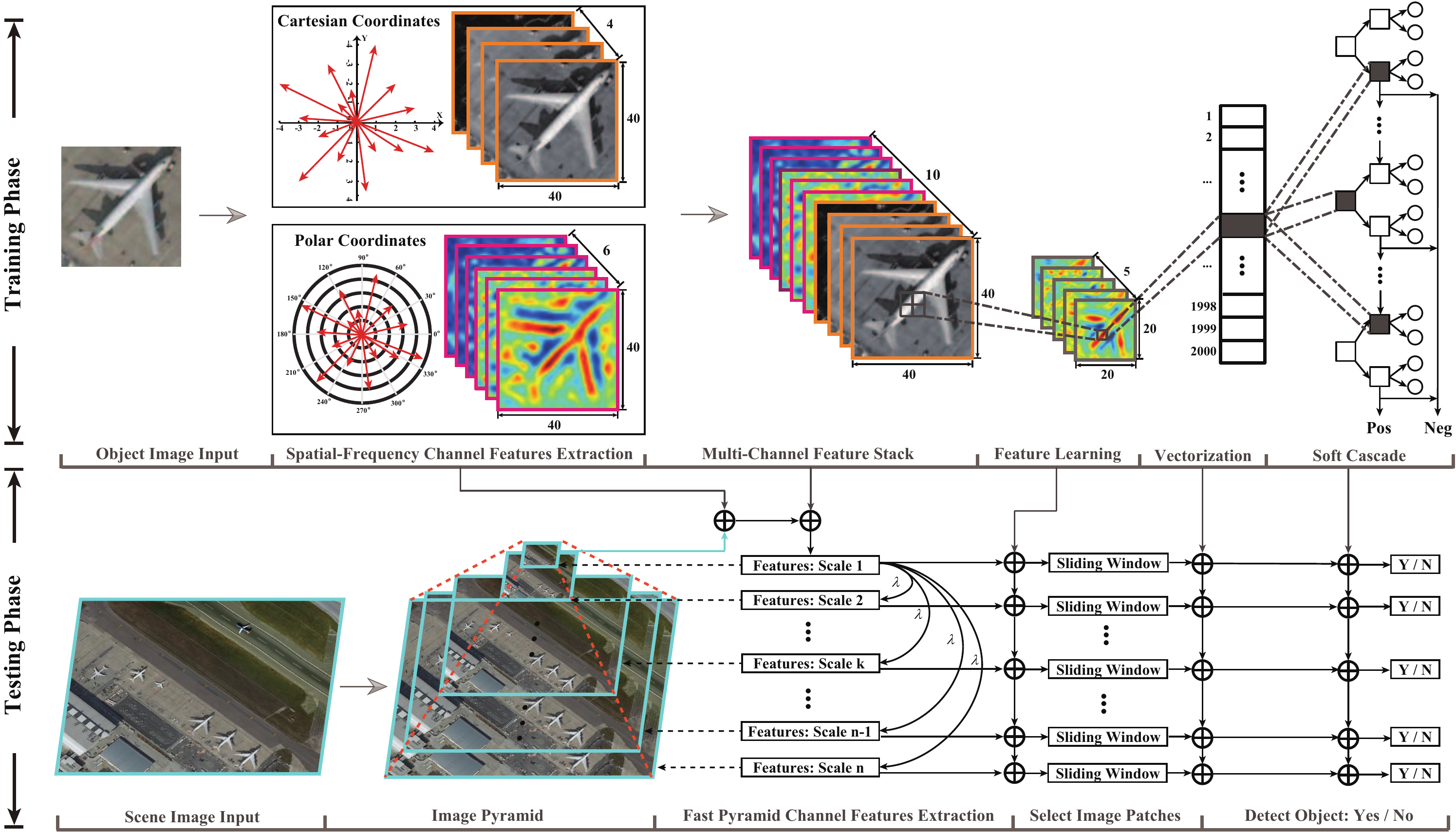}
\caption{The pipeline of ORSIm detector, which is a concatenation of rotation invariant descriptor with low-sampled image pyramid and boosting tree model learned with respect to diverse tasks.}
\label{fig2}
\end{figure*}

Many benchmarks available in public, e.g., TAS aerial car detection dataset \footnote{\url{http://ai.stanford.edu/~gaheitz/Research/TAS/tas.v0.tgz}}, NWPU VHR-10 dataset\footnote{The Vaihingen data was provided by the German Society for Photogrammetry, Remote Sensing and Geoinformation (DGPF).} \cite{Cheng2014,Cheng2016}, have contributed to spurring interest and progress in this area of remote sensing object detection. As the diversity of the database, many robust methods are born one after another in order to further improve the detection performances. Existing detection methods can be roughly categorized as follows \cite{cheng2016survey}: \textit{template matching-based},\textit{ knowledge-based}, \textit{object-based}, and \textit{machine learning-based} methods and other variants. These approaches mostly fail to describe object features in a complete space  with a densely set of scales. In our case, the so-called complete space should involve different properties robustly against various deformations, e.g., shift, rotation, etc. Moreover, a good image descriptor should be able to capture substantial image patterns with coarsely image pyramid. We will detail them close to our work and clarify the similarities and differences as well as pons and cons in the next section: \textit{Related Work}.
\subsection{Motivation and Objectives}
Object deformation (e.g., rotation, translation) in recognition or detection task is a common but still challenging problem. In particular, the remote sensing imagery is prone to have a more complex rotation behavior (see Fig. \ref{fig1}), due to its ``bird perspective''. Although the learning-based methods, such as deep neural networks (DNNs), deep convolutional neural networks (deep CNNs), have been proposed to learn the rotation-invariant features by manually augmenting the training set with different rotations, yet it is inevitably limited by the pre-setting rotation angles. This could be difficult to adaptively address the rotation problem of the fractional angle, thereby yielding a performance bottleneck. Another important factor that has a great effect on detection performance is the feature itself which can be manually designed or extracted by DNN. However, such powerful learning approaches fail to provide the richer representation without the strong support of large-scale labeled training samples.

Consequently, we mainly make our efforts to artificially develop or optimize the features towards the more discriminative rotation-invariant representations under the seminal object detection framework presented by Viola and Jones (VJ) \cite{Viola2004}, rather than the learning-based methods in this paper.
\subsection{Method Overview and Contributions}
To effectively address the aforementioned issues, the self-adaptive rotation-invariant channel features \cite{IJCV2014} are firstly constructed in polar coordinates, which has been theoretically proven to well fit the rotation of any angles. Furthermore, the shift-invariant channel features in Cartesian coordinates (e.g., color, gradient magnitude) are also extracted for the channel extensions in order to fully explore the potential of the feature representation, yielding a joint spatial-frequency channel feature (SFCF). We then step towards feature learning or refine (e.g., subspace learning, aggregated channel features (ACF)) to further refine the representations. Such features are finally fed into a boosting classifier with a series of depth-3 decision trees.

For the geospatial object detection in remote sensing, we propose a variant of VJ object detection framework, called \textbf{o}ptical \textbf{r}emote \textbf{s}ensing \textbf{im}agery detector (ORSIm detector). Unlike previous models in \cite{IJCV2014,PAMI2014} that are sensitive to translations and rotations, ORSIm detector is a more general and powerful framework robustly against various variabilities, particularly for remote sensing imagery. Additionally, a fast pyramid method is adopted to effectively investigate the multi-scaled objects without sacrificing the detection performance.  Fig. \ref{fig2} outlines the basic framework of ORSIm detector. The main highlights of our work are threefold.
\begin{itemize}
\item We propose a novel ORSIm detector by following the basic VJ framework by integrating spatial-frequency channel feature (SFCF), feature learning or refine, fast image pyramid estimation, and ensemble classifier learning (Adaboost \cite{freund1997decision});
\item A spatial-frequency channel feature is designed by simultaneously considering the invariance of rotation and shift in order to handle the complex object deformation behavior in remote sensing imagery;
\item An image pyramid generative model is simply but effectively embedded into the proposed framework by fast estimating a scaling factor in the image domain.
\end{itemize}

The remainder of this paper is organized as follows. Section II briefly reviews the previous work closely related to ours. Section III describes the proposed framework, including multiple domain feature exaction, feature stack, feature learning, training and testing. The experimental results on two datasets are reported in Section IV. Section V concludes our work and briefly discusses future work.
\begin{figure}[!t]
\centering
\includegraphics[width=0.95\linewidth]{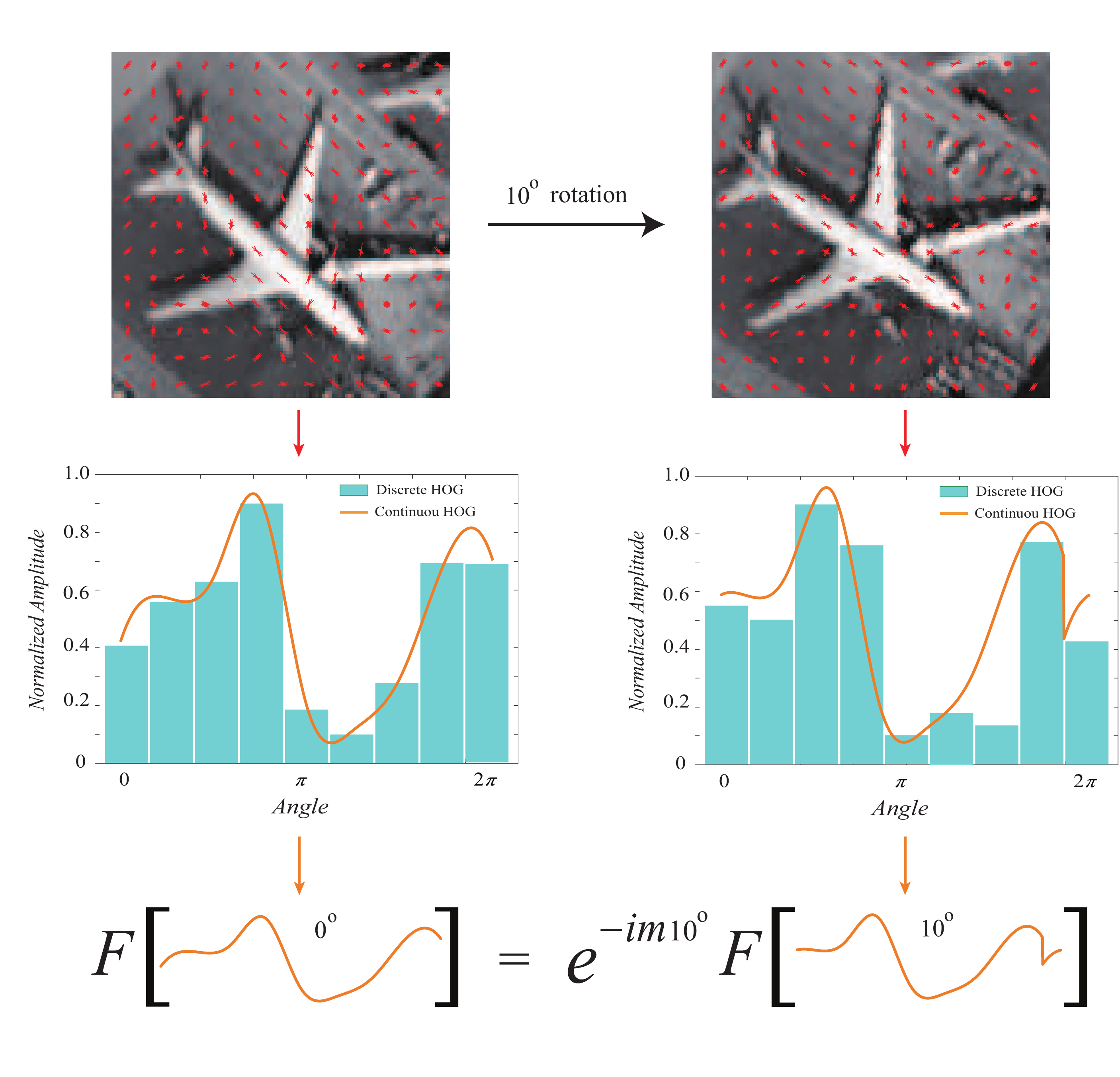}
\caption{ An Illustration of discrete and continuous HOG distribution function of a cell ($13 \times 13$ pixels). A reference HOG is shown on the left, and a ${10^ \circ }$ rotated HOG is given on the right. A property-rotated gap between the two discrete HOGs can be filled by shifting their corresponding continuous HOGs with $10^o$.}
\label{fig3}
\end{figure}
\section{Related Work}
In this section, several advanced techniques in object detection are introduced with the applications to remote sensing imagery. We also emphatically clarify our superiority, compared to three kinds of similar approaches partly associated with our work.
\subsection{Channel Features}
Channel Features refer to a collection of spatially discriminative features by linear or non-linear transformations of the input image. Over the past decades,  channel features extraction techniques have been received an increasing interest with successful applications in pedestrian detection \cite{PAMI2014,BMVC2009} and face detection \cite{IJCB2014,ICWMMN2015,ICCP2014}. Owing to their high representation ability, a variety of channel features have been widely used in geospatial object detection. Tuermer \textit{et al.} \cite{tuermer2013airborne} utilized the Histogram of Oriented Gradients (HOG) \cite{CVPR2005} as orientation channel features for airborne vehicle detection in a dense urban scene. Unfortunately, using orientation features alone is prone to hinder the detection performance from further improving. Inspired by the aggregate channel features (ACF) \cite{PAMI2014}, Zhao \textit{et al.} \cite{zhao2017effective} extended the channel features by additionally considering color channel features (e.g., gray-scale, RGB, HSV and LUV) to detect aircrafts through remote sensing images. However, these methods usually fail to achieve desirable performances due to the sensitivity to object rotation. For that, although many tentative works have been proposed to model the object's rotation behavior \cite{zhang2014object, wang2017feature}, yet the performance gain is still limited by the discrete spatial coordinate system.

With a theoretical guarantee, Liu \textit{et al.} \cite{IJCV2014} proposed a fourier histogram of oriented gradients (FourierHOG) with a rigorous mathematical proof. It models the rotation-invariant descriptor in a continuous frequency domain rather than in the discrete spatial domain using a Fourier-based convolutionally-manipulated tensor-valued transformation function $D = P(r){e^{im\varphi }}$. This function transfers the tensor-valued vectorized features (e.g., HOG \cite{hong2015novel}) to a scalar-valued representation, so as to make the features invariant with a maximized information gain. In contrast with HOG-like approaches that discretely compute the features (or descriptors) in the locally estimated coordinates from pose normalization, FourierHOG uses a smooth continuous function for fitting the statistical features in a continuous coordinate, as illustrated in Fig. \ref{fig3}. Furthermore, such a strategy can also avoid artifacts in the gradient binning and pose sampling of the HOG descriptor.

\begin{figure}[!t]
\centering
\subfigure[Training feature point]{\includegraphics[height=1.0in,width=1.65in]{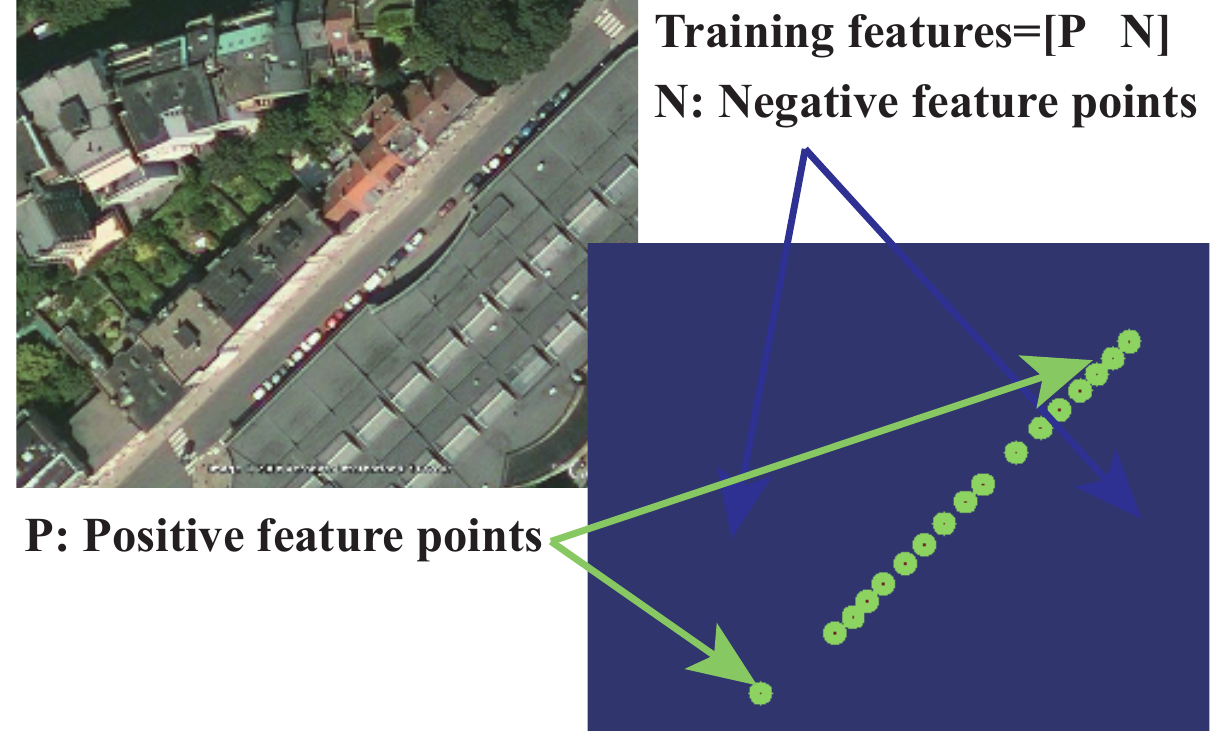}}
\subfigure[Feature channel cubic]{\includegraphics[height=1.0in,width=1.64in]{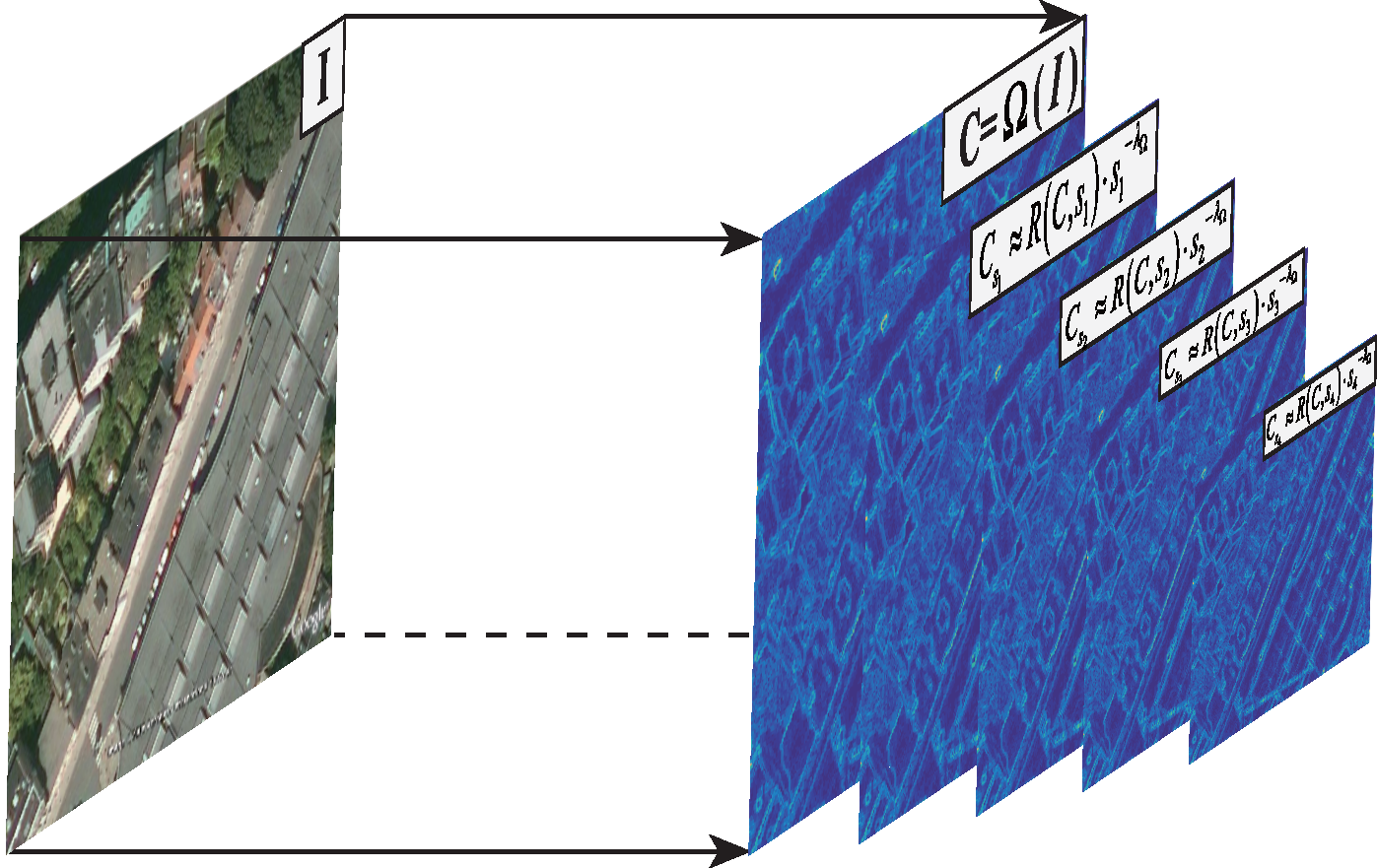}}
\caption{A simple sketch map of training feature points extracting in paper \cite{IJCV2014} and feature channel scaling in paper \cite{PAMI2014}.}
\label{fig4}
\end{figure}

Despite the superiority in representing rotation-invariance, FourierHOG ignores the importance of feature diversity. To this end, the proposed ORSIm extends the single channel features towards spatial-frequency joint ones, thereby further enriching the representations. On the other hand, FourierHOG, in fact, simplifies a challenging problem of object detection to that of object recognition. More specifically, the task of detecting boundary box of the object is converted into that of recognizing the central pixel to be either object or non-object, as illustrated in Fig. \ref{fig4}(a).
\subsection{Feature Channel Scaling}
Image multi-resolution decomposition is one of the essential techniques in high-level image analysis, such as object detection and tracking. The VJ framework is a seminal work for real-time object detection \cite{CVPR2001,Viola2004, IJCV2005}, which runs at $15$ frames per second (fps) for an image of $384\times 288$ pixels with a 700 MHz Intel Pentium III processor. Following this framework, HOG \cite{CVPR2005} yields a higher detection accuracy. Nevertheless, the two representative algorithms evenly sample scales in log-space and construct a feature pyramid for every scale. This is very time-consuming, and keeping the computational cost low is a significant challenge. Inspired by fractal statistics of natural images \cite{Ruderman1994}, Dollar \textit{et al.}\cite{PAMI2014} proposed a fast pyramid generative model by only estimating a scale factor, basically achieving a pyramid feature extraction in parallel. The key technique used in the model can be summarized as a feature channel scaling, as illustrated in Fig. \ref{fig4}(b), the goal of which is to compute finely sampled feature pyramids at a fraction of the cost by means of the fractal statistics of images. Furthermore, the features are computed at octave-spaced scale intervals in order to sufficiently approximate features on a finely-sampled pyramid. Therefore, these benefits make the model successfully applied to pedestrian detection at over 30 fps on an 8 cores machine Inter Core i7-870 PC \cite{benenson2012pedestrian}. Similarly, it has been also proven to be effective in aircrafts detection of remote sensing images \cite{zhao2017effective}. There is, however, an important assumption in the model, that is, the feature channels $\Omega$ are supposed to be any low-level shift-invariant in order to fit the operation of sliding windows, which makes the fast detection framework sensitive to angle variation or rotation-induced deformations. For this reason, Yang \textit{et. al.} \cite{IJCB2014} attempted to relax the constraint by learning varied face properties from multi-view images. The expensive cost of collecting multi-view remote sensing images still hinders Yang's algorithm from generalizing well. Congruously, either color channel features or FourierHOG is able to facilitate the use of the fast pyramid generative model, while their joint use (our SFCF) naturally does well.
\subsection{Boosting Decision Tree}
In the field of machine learning, the boosting methods have been widely used with great success for decades in various applications, e.g. object detection \cite{yokoya2015object,Ram2016,Zhang2016}, face detection \cite{Viola2004}, and pose detection \cite{Alb2016,nie2019hierarchical}. Unlike other powerful classifiers (e.g., Rotation-based SVM \cite{xia2016rotation}, structured SVM \cite{ICCV2011}, rotation forest \cite{xia2014hyperspectral}), the boosting-based ones iteratively select weak learners from a pool of candidate weak classifiers to deal with hard examples from the previous round, which can be treated as an enhanced model integrating former results and greedily minimizing an exponential loss function. Each weak learner is able to make the sample reweighed, then latter weak learners would more focus on those examples that are misclassified by former ones. Using this, a strong classifier can be learned with higher generalization ability and parameter adaptiveness.

The performance of boosting-based classifiers mainly relies on the discriminative ability of the feature and the number of weak classifier. In the next section, we will introduce the proposed unified framework (ORSIm detector) in semantically meaningful feature extraction, feature stack and learning as well as parameter selection of the boosting classifier.
\section{METHODOLOGY}
The proposed ORSIm detector starts with feature extractor. At this stage, spatial-frequency channel features are jointly extracted, including color and gradient magnitude channels from spatial domain and rotation invariant features from frequency domain. The features can be further refined by subspace learning or ACF, and then they can be fed into boosting decision tree for a better training and detection. \textbf{Algorithm 1} details the main procedures of the ORSIm detector.
\begin{algorithm}[!t]
{
\label{alg1}
\caption{ORSIm Detector}
\KwIn{Training data $Tr=[\mathbf{I}_{1},...,\mathbf{I}_{N}]$, and parameters.}
\KwOut{Model (detector), detection results}
 \textbf{Step 1:} \textit{Feature Extractor}\\
  \quad {\small 1) Extract pixel-wise spatial channel features by Eqs. (2-3);}\\
  \quad {\small 2) Extract pixel-wise frequency channel features by Eq. (10);}\\
  \quad {\small 3) Compute region-based SFCF representation by Eq. (11);}\\
  \textbf{Step 2:} \textit{Feature Learning or Refine}\\
  \quad {\small Perform a pooling-like operation and obtain ACF.}\\
  \qquad \qquad \qquad {\small ${\rm ACF} = Refine({\rm SFCF});$}\\
 \textbf{Step 3:} \textit{Training Ensemble Classifier}\\
 \While{$\varepsilon\rightarrow 0$}{
  \For{$t=1$ to $T$}{
       {\small $\mathcal{W}_{t} = TreeInitialization();$}\\
       {\small $\varepsilon_{t}=AdaBoost(\rm ACF, \mathcal{W}_{t});$}\\
       {\small $\beta_{t}=\varepsilon_{t}/(1-\varepsilon_{t});$}\\
       {\small $\mathcal{W}_{t}=UpdateWeights(\mathcal{W}_{t}, \beta_{t});$}\\
   }
   {$\varepsilon=\sum_{t=1}^{T}\varepsilon_{t};$}\\
  }
  \textbf{Step 4:} \textit{Test Phase with Feature Channel Scaling}\\
  \quad  {\small 1) Estimate the scale factor $\lambda$ by Eq. (12);}\\
  \quad  {\small 2) Obtain the feature pyramid of different scales;}\\
  \quad  {\small 3) Feed these features into the learned model (detector);}\\}
\end{algorithm}
\subsection{Spatial-Frequency Channel Features (SFCF)}
Commonly, the feature is limitedly represented in one single domain, this motivates the joint extraction of more discriminative features from the spatial and frequency domains to enrich the feature diversity.

Given a RGB remote sensing imagery $\mathbf{I}\in\mathbb{R}^{L \times W \times 3}$ as the input, we denote $F_{SFCF}$ as SFCF, mainly including the RGB channels, first-order gradient magnitude (GM) channel, and rotation-invariant (RI) channels, defined as
\begin{equation}
F_{SFCF} := \{ {\underbrace {\Omega _{1}(\mathbf{I})}_{\mathcal{RGB}},\;\underbrace {\Omega_{2}(\mathbf{I})}_{\mathcal{GM}},\;\underbrace {\Omega _{3}(\mathbf{I})}_{\mathcal{RI}}}\}
\end{equation}
where $\{\Omega_{i}(\mathbf{I})\}_{i=1}^{3}$ stands for the different feature sets.

1) \textbf{Pixel-wise Spatial Channel Feature:} In many tasks related to remote sensing, a color channel \cite{BMVC2010}, i.e. RGB, shows a strong ability in identifying certain materials sensitive to the color (e.g., tree, grass, soil, etc.), which can be denoted as
\begin{equation}
\Omega _{1}(\mathbf{I})=[F_{R}, F_{G}, F_{B}],
\end{equation}
where $F$ represents the channel features. Moreover, the normalized GM for the RGB image can be regarded as another important spatial channel features, since it can not only sharpen object edge, but also highlight small mutations that could be visually ignored in the smooth areas of the image, which has shown its effectiveness in detecting aerial or spaceborne objects \cite{tuermer2013airborne}. The resulting expression is
\begin{equation}
\Omega _{2}(\mathbf{I})=F_{GM}.
\end{equation}
2) \textbf{Pixel-wise Frequency Channel Feature:} The objects in remote sensing images, more often than not, suffer from various complex deformations. It should be noted that object rotation is one of the major factors that sharply leads to the performance degradation. Compared to extracting features in Cartesian coordinates, rotation invariance has been proven to more effectively analyze in Polar coordinates \cite{IJCV2014} where the feature can be separated as the angular information and radial basis $P(r)$, respectively. Let $\|\mathbf{d}\|$ and $\theta({\mathbf{d}})$ be the magnitude and the phase of a complex number $\mathbf{d}=dx+dyi$, where $dx$ and $dy$ are the horizontal and vertical gradients of a pixel in Cartesian coordinates, respectively. Coincidentally, the Fourier basis ${\psi_{k}}\left( \varphi  \right) = {e^{ik\varphi}} (k=0,1,...,m)$ is an optimal choice for modeling the angular part ($\theta({\mathbf{d}})$), theoretically proven in \cite{IJCV2014}, where $m$ stands for the Fourier order. The basis functions $\left[ {\psi {}_0,{\psi _1}, \cdots, {\psi _m} } \right]$ form harmonics on a circle, called circular harmonics. In \cite{IJCV2014}, the rotation behaviors $g(\bullet)$ in Fourier domain can be modeled by a multiplication or convolution operator. More specifically, given two $k$-th order Fourier representations in Polar coordinate ($f_{k_p}$ and $f_{k_q}$), then we have
\begin{equation}
\begin{array}{l}
g\left(f_{k_p} * f_{k_q} \right) = {e^{-i({k_p} + {k_q}){\alpha _g}}}\left[f_{k_p} * f_{k_q} \right] \circ \mathbf{T}_{g}\\
g\left(f_{k_p}f_{k_q} \right) = {e^{-i({k_p} + {k_q}){\alpha _g}}}\left[f_{k_p}f_{k_q} \right] \circ \mathbf{T}_{g},
\end{array}
\end{equation}
where $\mathbf{T}_{g}$ is a coordinate transform with a $\alpha_g$ relative rotation.

Given any one pixel ($\mathbf{p}$), its $k$-th order Fourier representations ($f_{k_{p}}$) can be further deduced by
\begin{equation}
\begin{array}{l}
f_{k_{p}} = \left\langle {h,{e^{ik_{p}\varphi }}} \right\rangle  = \frac{1}{{2\pi }}\int_0^{2\pi } {h\left( \varphi  \right){e^{ - ik_{p}\varphi }}} \\
\begin{array}{*{20}{c}}
{}&{ = \left\| \textbf{d}_{k_p} \right\|}
\end{array}{e^{ - ik_{p}\theta \left( \mathbf{d}_{k_p} \right)}},
\end{array}
\end{equation}
where $h(\varphi)$ is the distribution function of current pixel, which can be modeled by an impulse function with integral $\|\mathbf{d}_{k_p}\|$ \cite{IJCV2014} : $h(\varphi):=\|\mathbf{d}_{k_p}\|\delta(\varphi-\theta(\mathbf{d}_{k_p}))$.

When the Eq. (5) rotates by an angle $\alpha_g $, according to the rotation behavior $g\mathbf{d}: = {\mathbf{R}_g}\mathbf{d} \circ {\mathbf{T}_g}$ \cite{IJCV2014}, we have
\begin{equation}
\begin{array}{l}
{g}f_{k_{p}} = \left[{\|\mathbf{R}_{g}\mathbf{d}_{k_p}\|e^{-ik_{p}\theta(\mathbf{R}_{g}\mathbf{d}_{k_p})}}\right]  \circ {\mathbf{T}_g}\\
\begin{array}{*{20}{c}}
{}& =
\end{array}\left[{\|\mathbf{d}_{k_p}\|e^{-ik_{p}\alpha_{g}}e^{-ik_{p}\theta(\mathbf{d}_{k_p})}}\right]  \circ {\mathbf{T}_g}\\
\begin{array}{*{20}{c}}
{}&{ = {e^{ - ik_p{\alpha _g}}}\left[ {{f_{k_{p}}}\circ {\mathbf{T}_g}} \right]}.
\end{array}
\end{array}
\end{equation}

In order to make the feature rotation-invariant, namely $f_{k_p} = gf_{k_p}$,
we can set a set of filters (convolution kernels) with the same rotation behavior, denoted as $f_{k_q} (k=0,1,...,m)$. Using Eqs. (4-6), this can be formulated as
\begin{equation}
\begin{array}{l}
g\left(f_{k_p} * f_{k_q} \right) = {e^{-i({k_p} + {k_q}){\alpha _g}}}\left[f_{k_p} * f_{k_q} \right] \circ \mathbf{T}_{g},
\end{array}
\end{equation}
as long as satisfying $k_{p}+k_{q} = 0$, we can get
\begin{equation}
\begin{array}{l}
g\left(f_{k_p} * f_{k_q} \right) = \left[f_{k_p} * f_{k_q} \right] \circ \mathbf{T}_{g},
\end{array}
\end{equation}
thereby the convolutional features can be seen as the final rotation-invariant representation.

Inspired by the mentioned-above theory derivation in terms of rotation invariance, we construct the rotation-invariant features including the following three parts.
\begin{itemize}
\item Using the Fourier transformation on the input remote sensing images, the magnitude channel image in $k$-th Fourier order is naturally a kind of invariant feature, which is denoted as $F_{k_p}^{1}=||\mathbf{d}_{k_p}||$ ($k=0,1,...,m$) in a pixel-wise ($\mathbf{p}$) form.
\item To make the representation absolutely rotation-invariant, we get rid of rotation information from phase one by using Eqs. (7) and (8). That is, we generate a series of Fourier basis with equal and opposite order and use them on the Fourier representations of $\mathbf{I}$ ($f_{k_p}$) by a multiplication or convolution operation, which can be formulated as $F_{k_p}^{2}=f_{k_p}*f_{k_q}$, and $k_p=-k_q$.
\item We also consider a relative rotation-invariant feature representation by effectively utilizing the relative phase information \cite{giannakis1989signal}. Accordingly, this can be developed as a special rotation-invariant feature by coupling the convolutional features of two neighbouring kernel-radii (please refer to \cite{IJCV2014} for more details), which is formulated as $F_{k_p}^{3}=(f_{k_p}*f_{k_q, r1})\overline{(f_{k_p}*f_{k_q, r2})}/||(f_{k_p}*f_{k_q, r1})\overline{(f_{k_p}*f_{k_q, r2})}||$, and $k_p\neq -k_q$. $r1$ and $r2$ stand for the different convolutional kernels. 
\end{itemize}
\begin{figure}[!t]
\centering
\includegraphics[width=0.95\linewidth]{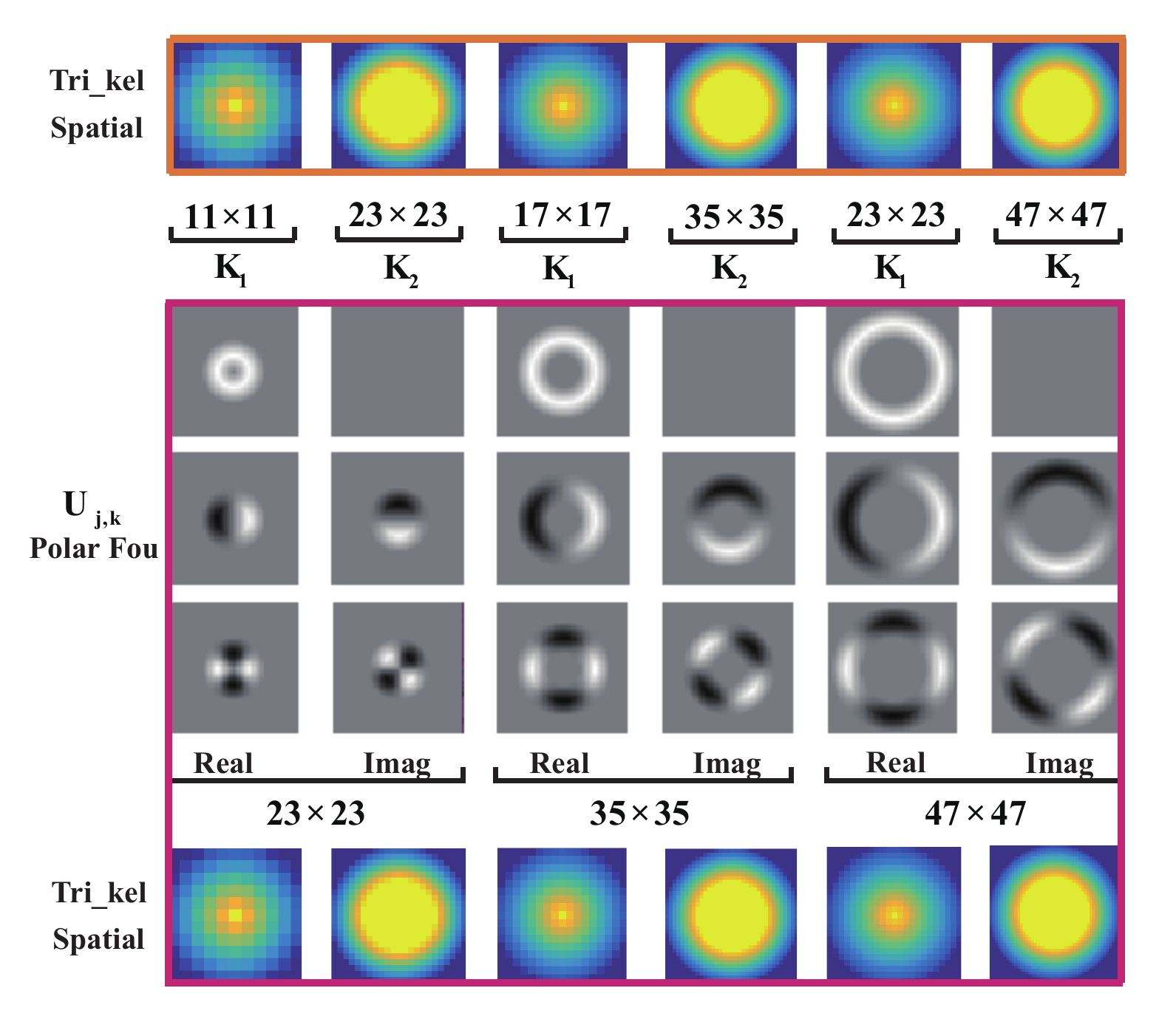}
\caption{Triangular convolution kernel of two domains. spatial ($K_1$: the convolution kernel for the spatial aggregation, $K_2$: the convolution kernel for the local normalization (based on gradient energy)); ${U_{j,k}}$: the basic function from triangular radial profile and a Fourier basis.}
\label{fig5}
\end{figure}

Therefore, the pixel-based frequency channel feature can be written by
\begin{equation}
\begin{aligned}
\Omega _{3}(\mathbf{p})&=\\
&[F_{0_p}^{1},...,F_{k_p}^{1},...,F_{0_p}^{2},...,F_{k_p}^{2},...,F_{0_p}^{3},...,F_{k_p}^{3},...],
\end{aligned}
\end{equation}
thus we have the image-level representation by collecting all pixel-based features
\begin{equation}
\Omega _{3}(\mathbf{I})=\{\Omega _{3}(\mathbf{p})\}_{p=1}^{L\times W}.
\end{equation}
3) \textbf{Region-based Channel Feature Representation:} Due to the low spatial resolution of remote sensing imagery, the detection performance is largely limited by the pixel-wise features. To better capture the semantically contextual information, we group pixel-wise channel features into region-based ones with kernel functions of different sizes. As visualized in  Fig. \ref{fig5}, we use the triangular convolution kernels, including isotropic triangles kernel, local normalization kernel, to extract region-based channel features in both spatial and frequency domains. Besides that, we additionally design a set of Fourier-based convolution kernels denoted as ${U_{j,k}} = {P_j}\left( r \right){e^{ik\varphi }}$ to construct the region-based rotation-invariant descriptors on the frequency domain (please refer to \cite{IJCV2014} about specific parameter settings of convolution kernels in details). Therefore, the resulting final SFCF is
\begin{equation}
\begin{aligned}
F_{SFCF}=[&\Omega _{1}(\mathbf{I})_{C_1},...,\Omega _{1}(\mathbf{I})_{C_j},...,\Omega _{2}(\mathbf{I})_{C_1},...,\\
&\Omega _{2}(\mathbf{I})_{C_j},...,\Omega _{3}(\mathbf{I})_{C_1},...,\Omega _{3}(\mathbf{I})_{C_j},...],
\end{aligned}
\end{equation}
where $\Omega _{i}(\mathbf{I})_{C_j}$ is the region-based features using the $j$-th convolution kernel.
\begin{figure}[!t]
\centering\includegraphics[width=0.95\linewidth]{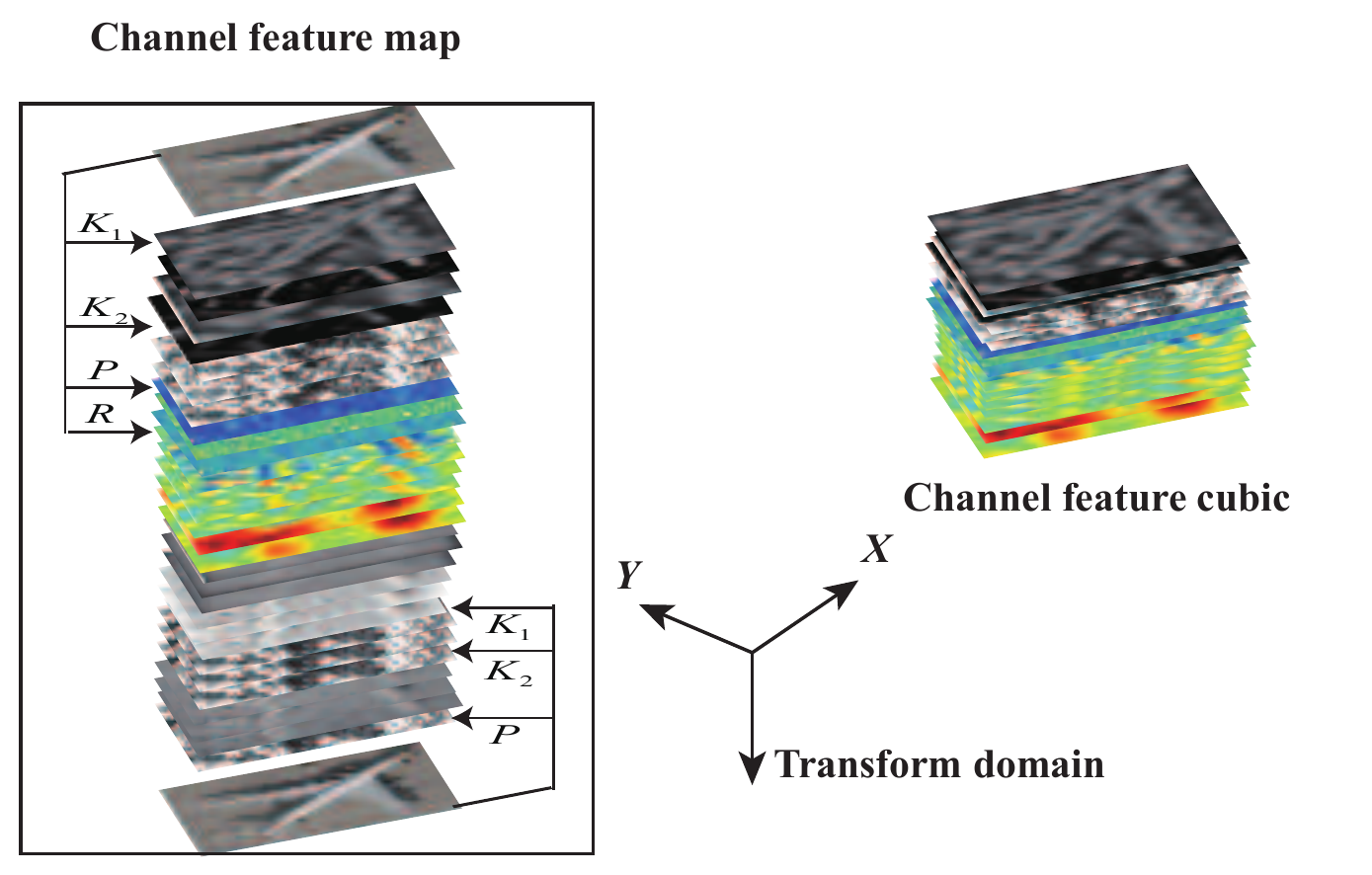}
\caption{Spatial-frequency channel features of an object sample and its region-based feature cubic.}
\label{fig6}
\end{figure}
\subsection{Feature Learning or Refine}
To effectively eliminate the feature gap between the two different domains and meanwhile improve its robustness and representative ability, we are able to learn or refine the feature cube (see Fig. \ref{fig6}) along spatial and channel directions using the following two strategies.

Module 1: Subspace-based learning (e.g., Principal Component Analysis (PCA) \cite{Roweis2000}). The extracted SFCF features can be further learned to reduce the computational and storage cost as well as improve the feature representation ability to some extent.

Module 2: Aggregation-based pooling. The SFCF can be also refined by the pooling-like operation (ACF) to dynamically adjust the support regions with different sizes and meanwhile maintain the structural consistence with the overall image \cite{IJCB2014}. Subsequently, the two-dimensional ACF is stretched to the one-dimensional fully-connected feature vector, making it better fitting into ensemble classifier learning. Inspired by the structurally encoding pattern, we select the ACF during the process of feature refine.

\subsection{Training Phase with Ensemble Classifier Learning}
Up to the present, boosting is one of the most popular learning techniques by integrating a large number of weak learners to generate a stronger one. The boosting-based method (e.g., AdaBoost) is built on the fact that those selected week classifiers should minimize the training errors and keep or reduce the test errors. For this reason, we apply a soft-cascade boosting structure with the depth-3 decision trees\cite{PAMI2014}, which is capability of discriminating intra- and inter-samples more effectively and simultaneously playing a role in feature selection. Significantly, the learning strategy is robust against background interference in object detection, especially in more complex scene of remote sensing imagery.
\subsection{Test Phase with Feature Channel Scaling}
Sliding window is a commonly used detection technique in testing phase behind extracting finely-sampled image pyramid. However, it implies a heavy computational cost, which is not a good tool in the real-world. A fast image pyramid model \cite{PAMI2014} introduced in Section II.B is implemented in our framework by automatically estimating scaling factor of feature channels, which is expressed as
\begin{equation}
{\mathbf{C}(\mathbf{I},s)} \approx \Omega(\mathbf{R}(\mathbf{I},s))= \mathbf{R}(\mathbf{I},s) \cdot {s^{ - {\lambda _\Omega }}},
\end{equation}
where $\mathbf{I}$ is an input image, and $\mathbf{R}(\mathbf{I},s)$ is a re-sampled image of $\mathbf{I}$ by $s$. $\mathbf{\lambda}$ is a scaling factor to be estimated. The corresponding channel image at a scale $s$ can be presented by Eq. (12). The different channels can be computed with a linear or a non-linear transformation of the original image in the spatial and frequency domains. Using Eq. (12), we can quickly obtain the channels features of all pyramid images using the given $\lambda$ calculated in the training phase.
\section{Experimental Results and Analysis}
\subsection{Optical Remote Sensing Datasets}
In this section, two well-known public optical remote datasets: car targets in satellite dataset \footnote{\url{http://ai.stanford.edu/~gaheitz/Research/TAS/tas.v0.tgz}} and airplane targets in NWPU VHR-airplane dataset \footnote{\url{http://www.ifp.uni-stuttgart.de/dgpf/DKEPAllg.html.}} are used to quantitatively evaluate the performances of the proposed method. In our work, 60\% samples are assigned as training set and the rest is testing set for both datasets. The main focus of this paper is to create a more robust and discriminative feature representation, ensuring rotation and translation invariance. Generally, it is very expensive and time-consuming to collect a large number of training samples, particularly labeling remote sensing data. Therefore, it is very meaningful and challenging for users to assess the generalization performance of the classifier with a limited training set. To stably evaluate the performance of the proposed method, we conduct 5-fold cross-validation and report an average result below across the folds.
\begin{figure}[!t]
\centering
\subfigure[Satellite dataset]
		  {\includegraphics[width=0.24\textwidth]{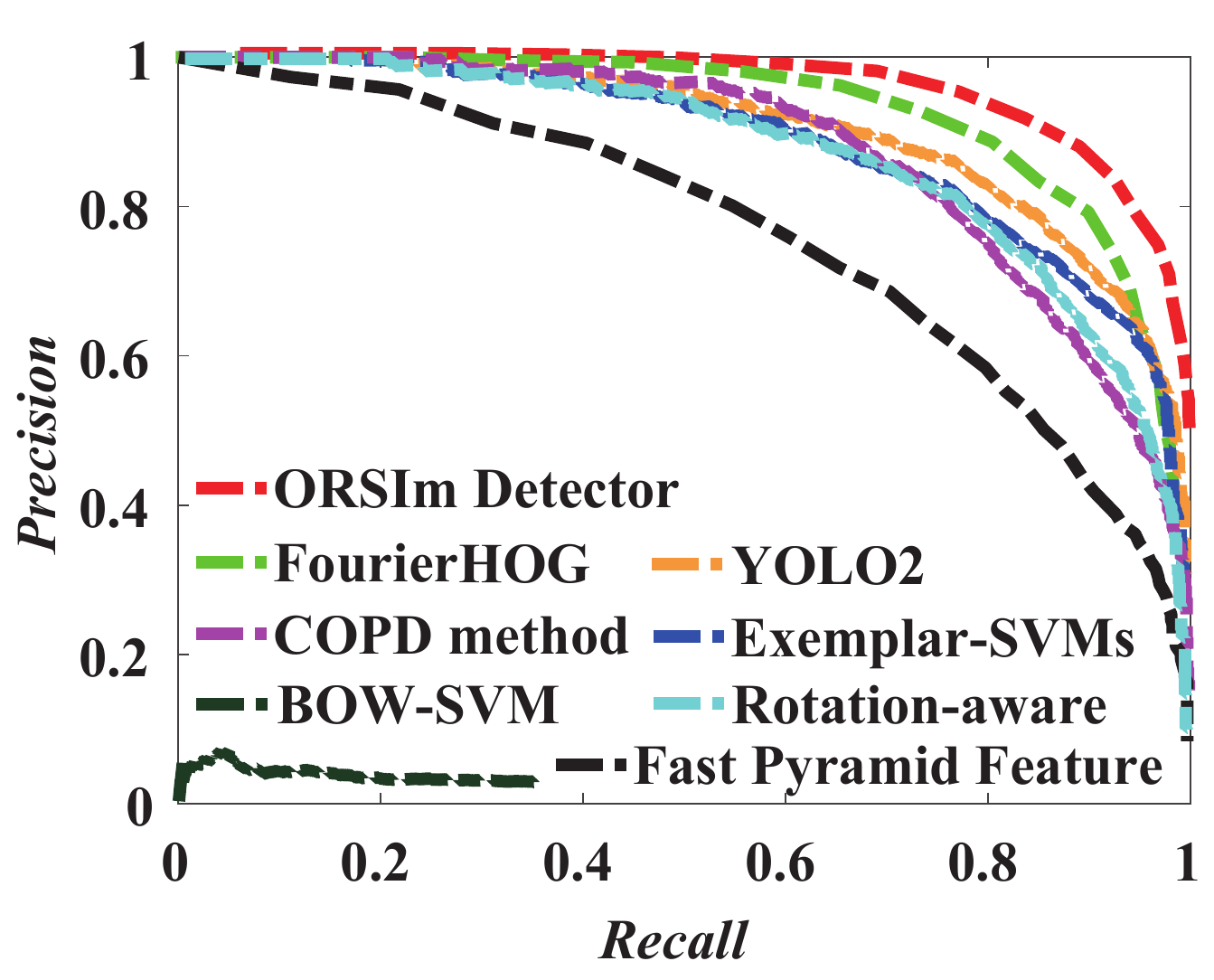}}
\subfigure[NWPU VHR-Airplane]
		  {\includegraphics[width=0.24\textwidth]{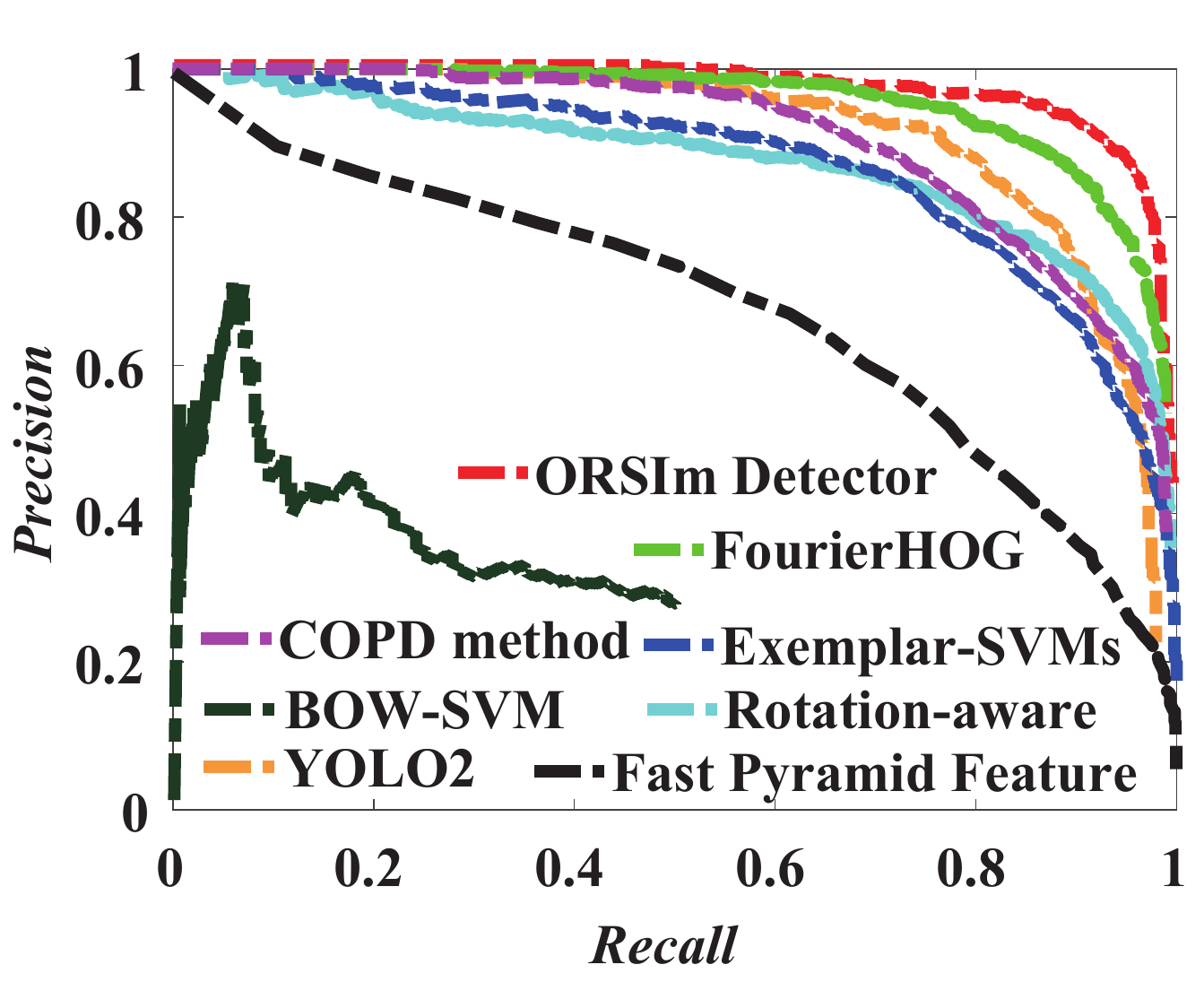}}
\caption{PR curves of the proposed ORSIm detector in comparison with state-of-the-art approaches.}
\label{tab:PRCurve}
\end{figure}

1) Satellite dataset: This dataset was acquired from Google Earth \cite{ECCV2008}. The low resolution and the varying illumination conditions caused by the shadows of buildings make this dataset very challenging. In detail, the images contain $1319$ manually labeled cars from 30 images with the size of $792 \times 636$. At the training stage, all car windows are rescaled to $40 \times 40$, due to the average car window is approximately this window. Also, their mirror images are used to double the positive images as data augmentation in all experiments, which can avoid over-fitting and improve generalization ability. Meanwhile, the negative images are cropped at random positions from the $226$ natural images without any car objects.

2) NWPU VHR-10 dataset: This dataset consists of 10 different object detection datasets acquired from Google Earth (spatial resolution 0.5m-2m) and Vaihingen data set (spatial resolution 0.08m) \footnote{The Vaihingen data was provided by the German Society for Photogrammetry, Remote Sensing and Geoinformation (DGPF).} (Please refer to \cite{Cheng2014,Cheng2016} for more details). To meet our experimental assumption, that is, we mainly aim at detecting those objects with highly rotation behavior, hence airplane is proper research objects, which are selected as our another experimental data to effectively evaluate our method. More specifically, the positive image set without any outliers is composed of 650 airplane images and each of them includes at least one target. The negative image set consists of 150 images without any class-relevant targets. The original maximal and minimal windows are set to $130 \times 120$ and $40 \times 40$ pixels, respectively. Additionally, the number of positive images in training set is doubled by mirror processing, while the negative images are randomly selected from the 100 images without any airplanes.
\subsection{Experimental Setup}
All the experiments in this paper were implemented with Matlab2016 on a Windows 7 operation system and conducted on an Intel Xeon 2.6GHz PC (CPU) with 128GB memory. Morevoer, there are several important modules in the proposed ORSIm framework, such as SFCF extraction, sampling window, smoothing, feature pyramid, and classifier setting. We will gradually detail them in the following.

\textbf{SFCF extraction:} The channel features used in our case mainly consist of two parts: spatial channel features and frequency channel features. The former involves color channels and corresponding magnitude of gradient channels, and the latter is the rotation-invariant feature channels. More specifically, RGB (red, green, blue), LUV (luminance, chromaticity coordinates) and HSV (hue, saturation, value) are selected as the potential color spaces. The magnitude of gradient channel is set as the magnitude of the channel with the maximal gradient amplitude response. There are three parts in the rotation-invariant channels, which are the true rotation invariant features (same Fourier orders, e.g. $m_1+m_2=0$), the magnitude features, and the coupling features across different radius (please refer to \cite{IJCV2014} for more details). During the process, two parameters need to be considered, namely the radii ($r$) of convolutional kernels and the number of Fourier order ($m$). We assign five scales with six different half-width of $\sigma=\left\{ {3,4,5,6,7,8} \right\}$ to the value of $r$, i.e. $\sigma = 6, {r_j} \in \left\{ {0,6,12,18,24} \right\}$, while the $m$ is set to $2,3,4,5$, as suggested in \cite{IJCV2014}.

\textbf{Sampling window:} Due to the fact that objects in a scene hold the different resolution, it is necessary for objects (e.g., vehicle, airplane) to be upsampled or downsampled to a consistent size. Therefore, we attempt to search an optimal length-width ratio in a proper range, by resizing the cars on the satellite dataset to $28 \times 24$, $32 \times 28$, $40 \times 36$, and $44 \times 40$, as well as the airplanes on NWPU VHR-airplane dataset to $56 \times 56$, $64 \times 64$, $72 \times 72$, $80 \times 80$, and $88 \times 88$.

\textbf{Smoothing:} The smoothing operation has been proven to be effective in improving the representation ability of the features \cite{IJCB2014,PAMI2014}. Similarly, we perform smoothing before feature computation (pre-smoothing) and after feature learning or refine (post-smoothing) with the binomial filter. The filter radius is set to 1 in our setting.

\textbf{Feature pyramid:} The fast feature pyramid in \cite{PAMI2014} is applied in the proposed ORSIm framework by coarsely sampling feature channels in order to speed up the hard negative mining and test phase without additional loss of detection precision. We sample the objects in the four different scales ($s$= 1, 2, 4, 8) with the sampling rate of $2^{ -\frac{1}{nPerOct}}$. The smallest pyramid image is determined by the size of sampling window, and the largest one has the same size as the original image.
\begin{figure*}[!t]
\centering
\centering\includegraphics[width=0.95\linewidth]{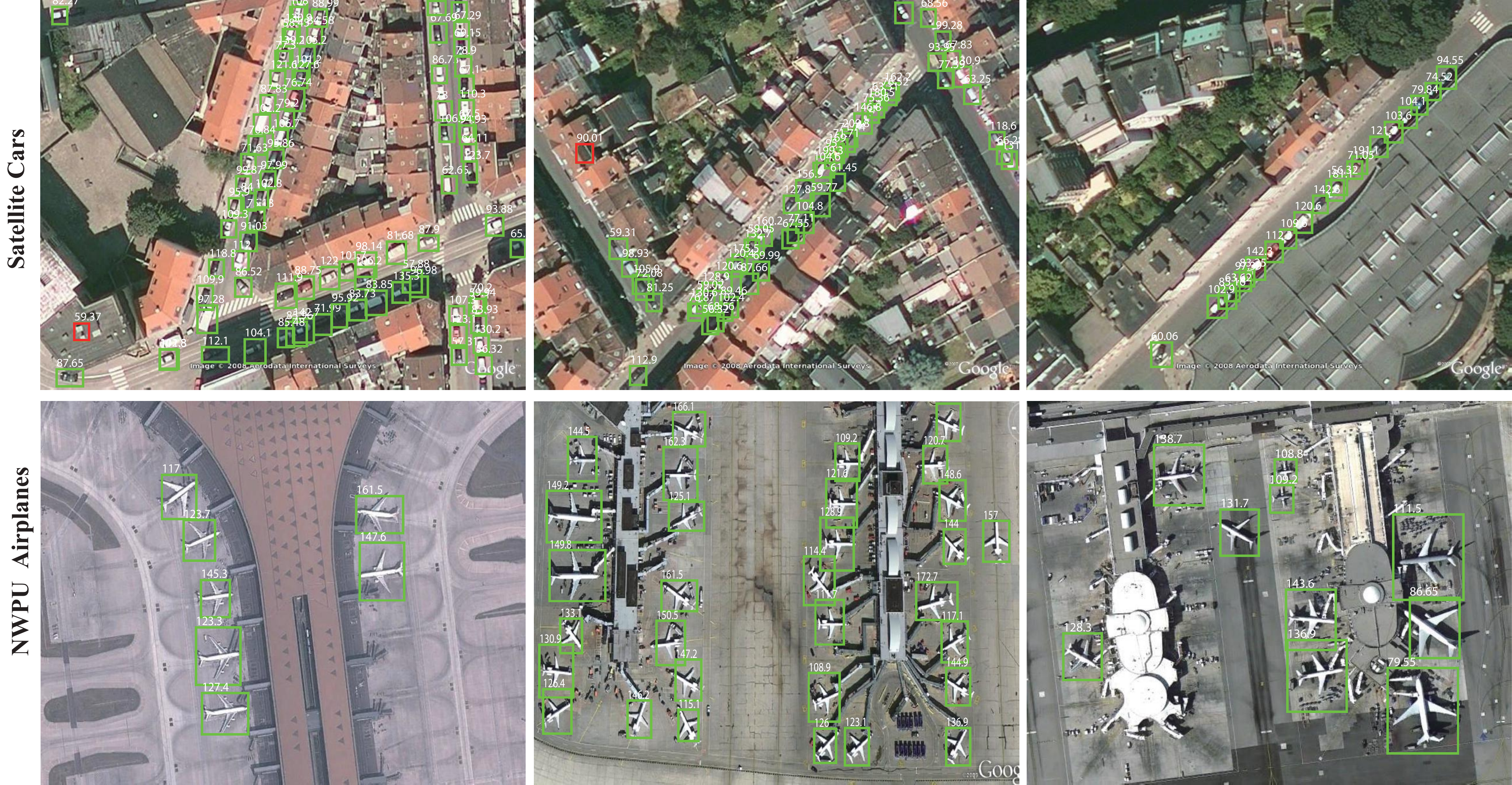}
\caption{Some visual detection results (false detection in red, true positive in green, and missing detection in blue) by using the proposed method on the two different datasets.}
\label{fig:visual result}
\end{figure*}

\textbf{Classifier setting:} AdaBoost \cite{TAS2000}, which is a boosting-based ensemble classifier learning, is used to train the classifier. To train a stronger learner, we use a weighted majority voting to generate the boosting decision tree by combining the hypotheses obtained from those diversified weaker learners. To avoid over-fitting, we gradually increase the number of weak learners from 32 to 2048. It is worth noting that negative samples used in training phase and testing phase are selected using a sliding window and a coarsely sampled image pyramid instead of point-based operators as presented in \cite{IJCV2014}.

\textbf{Evaluation criteria:} Four criteria, Precision-Recall (PR) curve, Average Precision (AP), Average Recall (AR), and Average F1-score (AF), are adopted to quantitatively evaluate the detection performances. More precisely, when the rate between the overlap of the detection bounding box and the ground-truth box exceeds $50\%$, it is counted as a true positive (TP); otherwise, as a false negatives (FN). Therefore, the final Precision (P) is computed by $\frac{TP}{TP+FP}$, and the Recall (R) is $\frac{TP}{TP+FN}$, while F1-score can be computed by $F1 = \frac{2\times P\times R}{P+R}$. AP is used as a global indicator to assess the performances of the algorithm.

\subsection{Experimental Results}
\begin{figure*}[!t]
\centering
\subfigure[Original NMS]
		  {\includegraphics[width=0.47\textwidth]{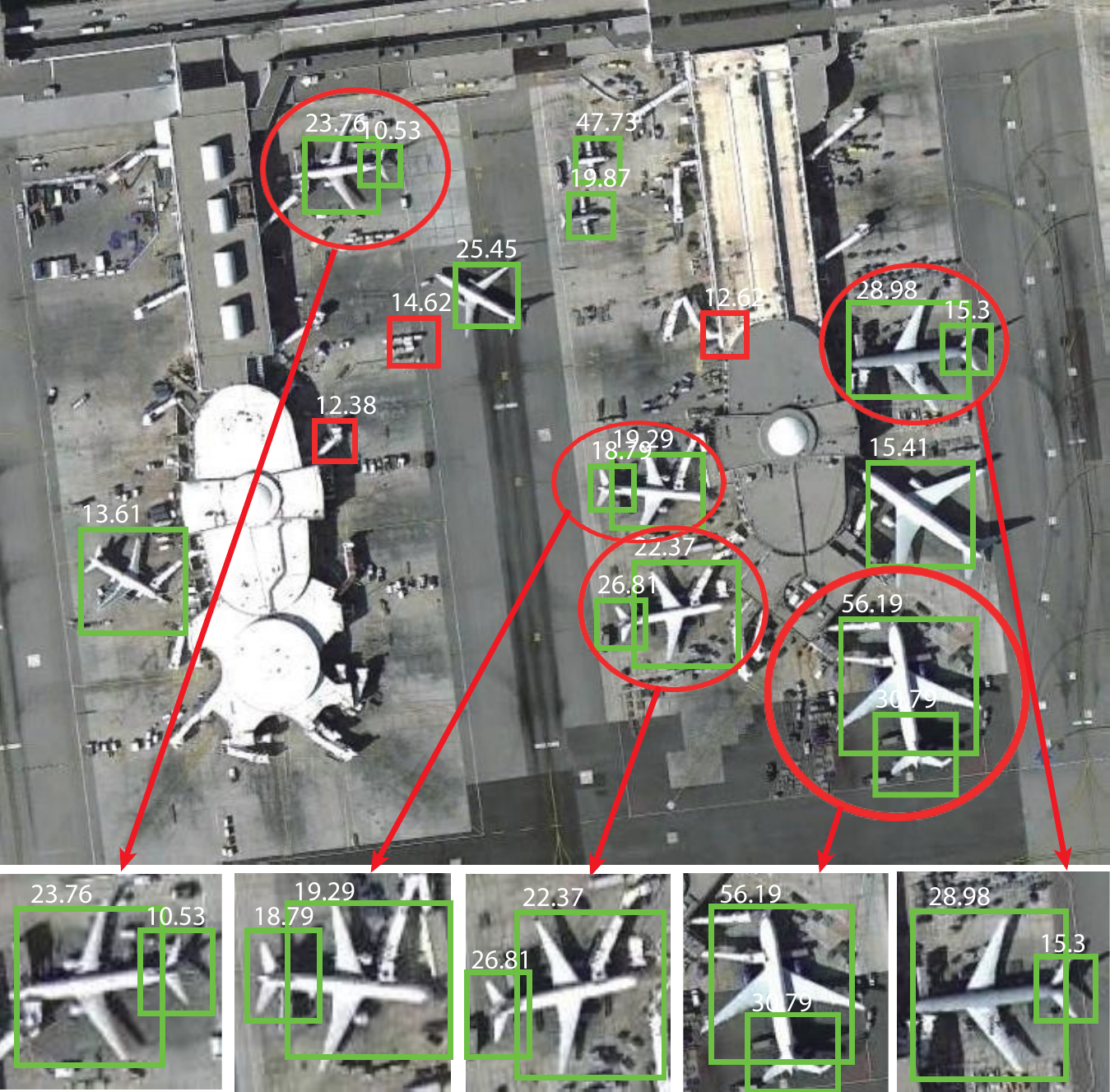}}
\subfigure[Two-step NMS]
		  {\includegraphics[width=0.47\textwidth]{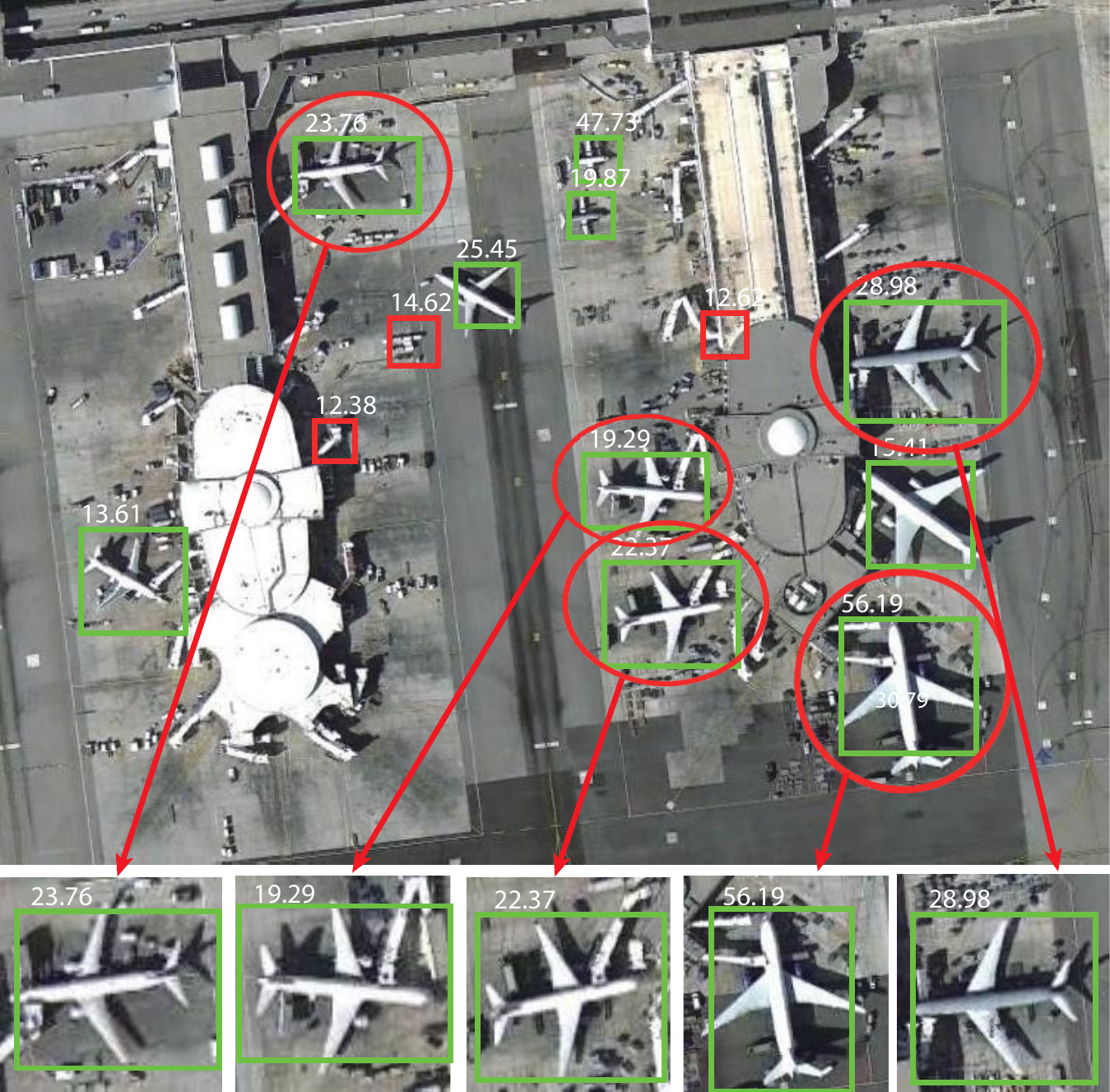}}
\caption{Overlap removal using original NMS and two-step NMS.}
\label{fig:NMS}
\end{figure*}
\begin{table*}[!t]
  \centering
  \caption{Performance comparisons of eight different methods in terms of AP, AR, and AF. The best results are shown in bold.}
    \begin{tabular}{l||c||c|c|c|c|c|c|c|c}
    \hlinew{1.5pt}
    \multirow{2}{*}{Method}&
    \multirow{2}{*}{Image pyramid}&\multicolumn{4}{c|}{Satellite (\%)}&\multicolumn{4}{c}{NWPU VHR-Airplane (\%)}\\
    \cline{3-10} & & AR & AF & AP & FPS & AR & AF & AP & FPS \\ \hline\hline
    Exemplar-SVMs & Standard pyramid & 78.29 &81.62 & 85.25 & 0.92 & 80.28 & 80.32 & 80.37& 0.67 \\
    Rotation-aware  & Standard pyramid & 77.64 & 78.80 & 80.01 & 1.28 & 81.76 & 80.54 & 79.35 & 0.61 \\
    COPD-based & Standard pyramid & 76.21 & 78.23 & 80.37 & 1.05 & 75.17 & 79.84 & 85.13 & 1.06 \\
    BOW-SVM & Standard pyramid & 8.29 & 10.77 & 15.38 & 1.16 & 3.58 & 6.27 & 25.12 & 1.17 \\
    YOLO2 (GPU) & --- & 83.25 & 85.70 & 88.30 & 9.12 & 84.92 & 87.27 & 89.75 & 9.13 \\
    FourierHOG & Standard pyramid & 88.20 & 89.78 & 91.42 & 0.83 & 87.78 & 88.97 & 90.20 & 0.66 \\
    ACF & fast pyramid & 62.14 & 68.59 & 76.53 & 8.98 & 62.56 & 64.04 & 65.59 & 8.05 \\
    ORSIm Detector & fast pyramid & \bf 91.26 & \bf 93.01 & \bf 94.83 & 4.94 & \bf 91.12 & \bf 93.21 & \bf 95.39 & 4.72 \\
    \hlinew{1.5pt}
    \end{tabular}
    \label{tab:eight approach}
\end{table*}

\begin{table*}[!t]
\centering
\caption{Performance comparisons of three different classifiers. The best results are shown in bold.}
    \begin{tabular}{l||c|c|c|c|c}
    \hlinew{1.5pt}
     Dataset& Method & HOG \cite{CVPR2005} & ACF \cite{BMVC2009,PAMI2014} & FourierHOG & our SFCF\\ \hline \hline
    \multirow{3}{*}{Satellite} & Linear SVM & 65.30 & 68.72 & 83.19 & 85.18\\
    & RF & 71.20 & 73.12 & 86.77 & 88.29 \\
    & AdaBoost & 74.38 & 76.53 & 90.42 & \bf 93.12 \\ \hline
    \multirow{3}{*}{NWPU VHR-Airplane} & Linear SVM & 74.38 & 76.67 & 80.01 & 85.12\\
    & RF & 70.05 & 74.98 & 87.26 & 91.68 \\
    & AdaBoost & 77.98 & 80.37 & 90.20 & \bf 94.31 \\
    \hlinew{1.5pt}
    \end{tabular}
\label{tab:performance_comparison}
\end{table*}
\begin{figure*}[!t]
\centering\includegraphics[width=0.95\linewidth]{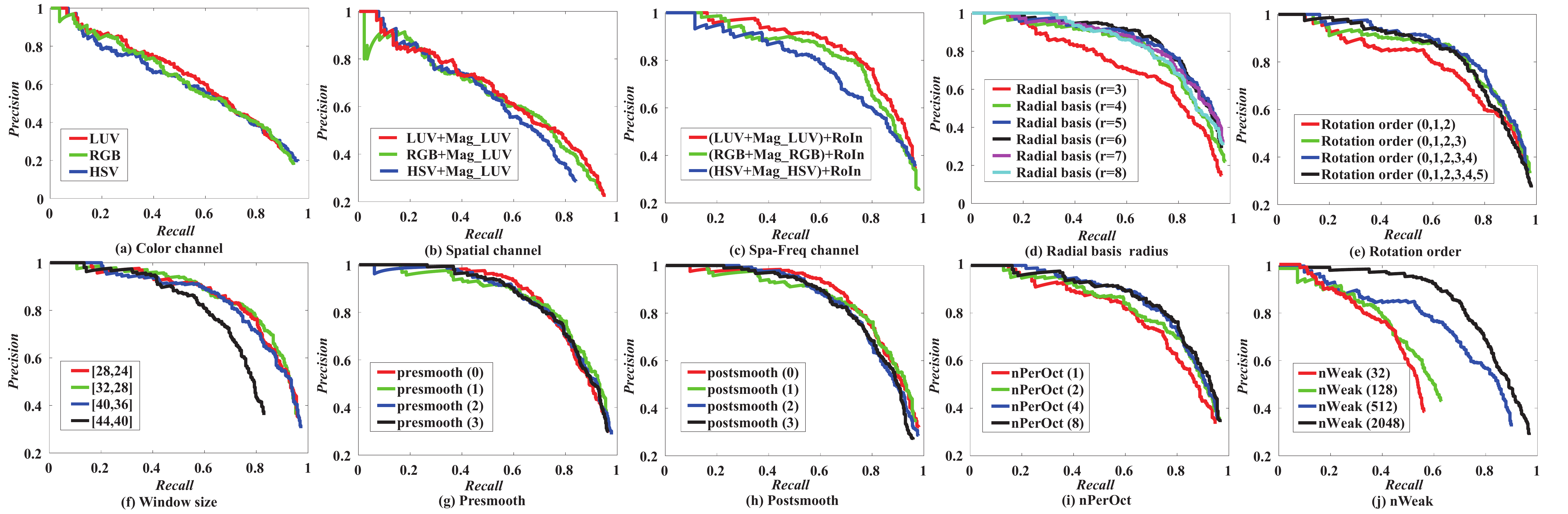}
\caption{Performance comparison of ORSIm detector under different parameter setting on the satellite dataset}
\label{tab:carparam}
\end{figure*}
\begin{figure*}[!t]
\centering\includegraphics[width=0.95\linewidth]{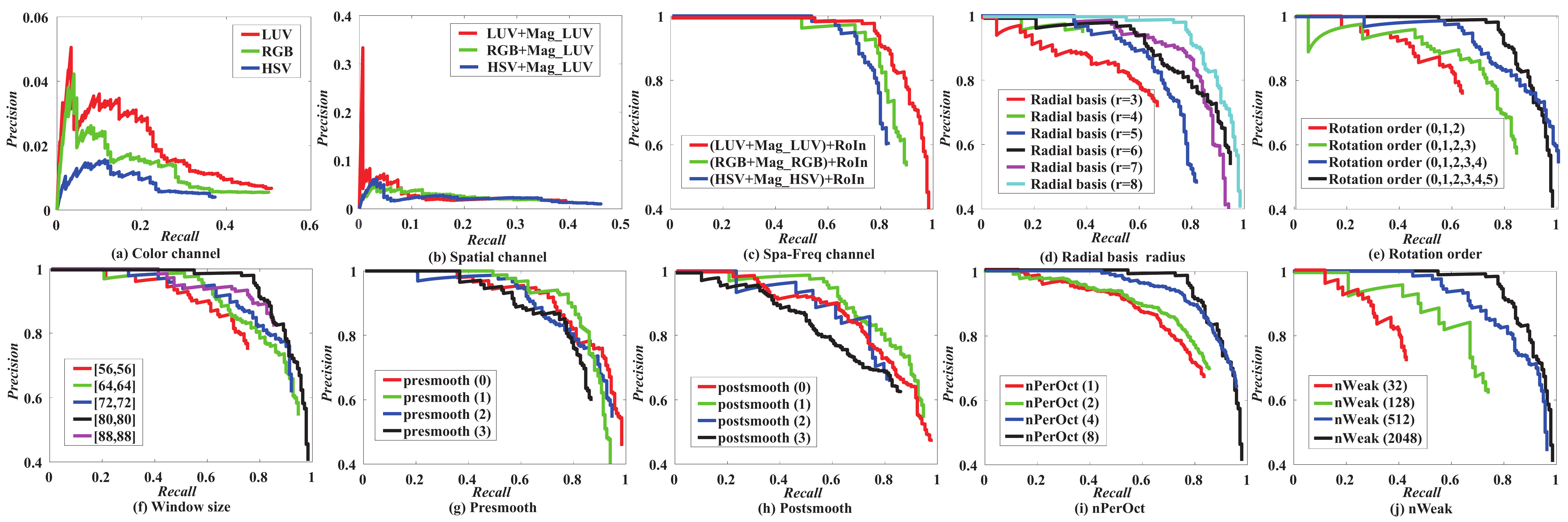}
\caption{Performance comparison of ORSIm detector under different parameter setting on theNWPU VHR-Airplane dataset}
\label{tab:airparam}
\end{figure*}
\subsubsection{Discussion on classifier selection}
As listed in Table \ref{tab:performance_comparison}, those methods based on support vector machines (SVMs) or random forest (RF) classifier also achieve the good performances. This motivates us to have a great interest in investigating the classifier selection. To this end, three different classifiers (e.g., linear SVM, RF, and AdaBoost\footnote{AdaBoost \cite{TAS2000}, also known as AdaBoost-DTree, is used in our framework.}) are used to evaluate the detection performance under four different feature descriptors, that is, HOG, ACF, FourierHOG, and our SFCF, as detailed in Table \ref{tab:performance_comparison}. For a fair comparison, the parameters used in the three classifiers are optimally tuned by cross-validation on the training set. Overall, the linear SVM yields the relatively poor performances, compared to the results of RF. Because the RF is more robust than linear SVM to some extent, especially when the training samples are limited. Furthermore, the AdaBoost performs better than the two other classifiers. Two possible factors could explain the results. On one hand, AdaBoost is a boosting-based ensemble classifier learning, which can generate a more robust strong classifier by weighing a large number of weak classifiers. Consequently, it holds a more powerful performance than the linear SVM in recognition and classification. On the other hand, although the RF and AdaBoost are both based on the boosting-like strategy, yet the RF equally puts the weights on each sub-classifier and the AdaBoost adaptively weighs each weak classifier by iteratively updating weights. This makes the resulting final classifier generated by Adaboost more suitable for the current dataset, thereby yielding a better performance.
\subsubsection{Overview of performance comparison}
To quantitatively assess the detection performances of the proposed method, we compare several state-of-the-art methods related to our framework, such as Exemplar-SVMs \cite{Malisiewicz2011}, rotation-aware features \cite{Schmidt2012}, COPD-based \cite{Cheng2014}, BOW-SVM \cite{BOW2010}, fast feature pyramids \cite{PAMI2014}, You Only Look Once (YOLO2) \cite{YOLO2017}\footnote{Similarly to \cite{Cheng2016} and \cite{wu2018msri}, data augmentation by the rotation and translation of the training samples are performed.}, FourierHOG \cite{IJCV2014}\footnote{We select positive and negative samples by sliding windows rather than points for a fair comparison.}. Fig. \ref{tab:PRCurve} shows the PR curves of different algorithms on the two datasets and Table \ref{tab:eight approach} correspondingly lists the quantitative results in terms of average precisions and mean running times. Accordingly, we can make the following observations. The Exemplar-SVMs and Rotation-aware methods have similar performances, as the standard HOG features and discrete grid sampling are used. Not surprisingly, BOW-SVM and ACF yield the worst performances because they ignore the spatial contextual relationships among the local features and are limited by the rotation-related representation ability. Although the detection performance might be improved by modeling a deeper network and embedding anchor boxes, yet YOLO2 is not robust to tiny object and arbitrary pairs of objects that are not more than a tiny distance apart. FourierHOG holds a slightly lower performance than ours but much better than others on the two datasets, which indicates that the point-based feature representation is insensitive to resolution. As expected, the proposed ORSIm detector largely outperforms the other investigated methods on both datasets, which shows its effectiveness and superiority. This also can be demonstrated from Table \ref{tab:performance_comparison} that the precision of ORSIm detector is dramatically higher than that of the others owing to the well-designed SFCF and the use of AdaBoost. It is worth noting in Table \ref{tab:eight approach} that the methods with fast feature pyramid allow for faster detection than those without it. Despite of slowing down the speed (relatively lower than ACF and YOLO2\footnote{The code is run on the tensorflow using GPU, which is available from the website: https://github.com/simo23/tinyYOLOv2.}), the proposed ORSIm detector acquires the highest detection precision.

Visually, little roofs are wrongly identified as cars, and there are also some leak detection in transport cars, as shown in the first row of Fig. \ref{fig:visual result}. This might result from a limited number of training samples and unbalanced class distribution. In addition, a weaker visible edge might mislead the classifier, since the transport cars are white. Compared to car detection in a complex urban scene, false detection of the airplanes also occurs when background and targets have similar shape and color, i.e. the tail of the airplane (see Fig. \ref{fig:NMS}(a)). But this issue can be well fixed by a two-step nonmaximum suppression (NMS) algorithm \cite{zhao2017effective}. The improved results can be found in Fig. \ref{fig:NMS}(b). 
\subsection{Sensitivity Analysis}
We experimentally analyze and discuss the potential influences under the different configuration of the proposed ORSIm detector, making it possible to generalize well in more datasets. The optimal combination is finally determined by 5-fold cross-validation on the training set.
\subsubsection{Towards Parameter Setting}
Figs. \ref{tab:carparam} and \ref{tab:airparam} show the performance comparison of the different parameter setting on the two used datasets. More specifically, the LUV color space performs better than the two others on both dataset, and even more obvious when using a combination of the color channels with gradient magnitude channel. Interestingly, there is a similar trend after adding the rotation-invariant feature channels, as shown in Figs. \ref{tab:carparam} and \ref{tab:airparam} (c).
\begin{figure}[!t]
\centering
\subfigure[Vehicle dataset]
          {\includegraphics[width=0.24\textwidth]{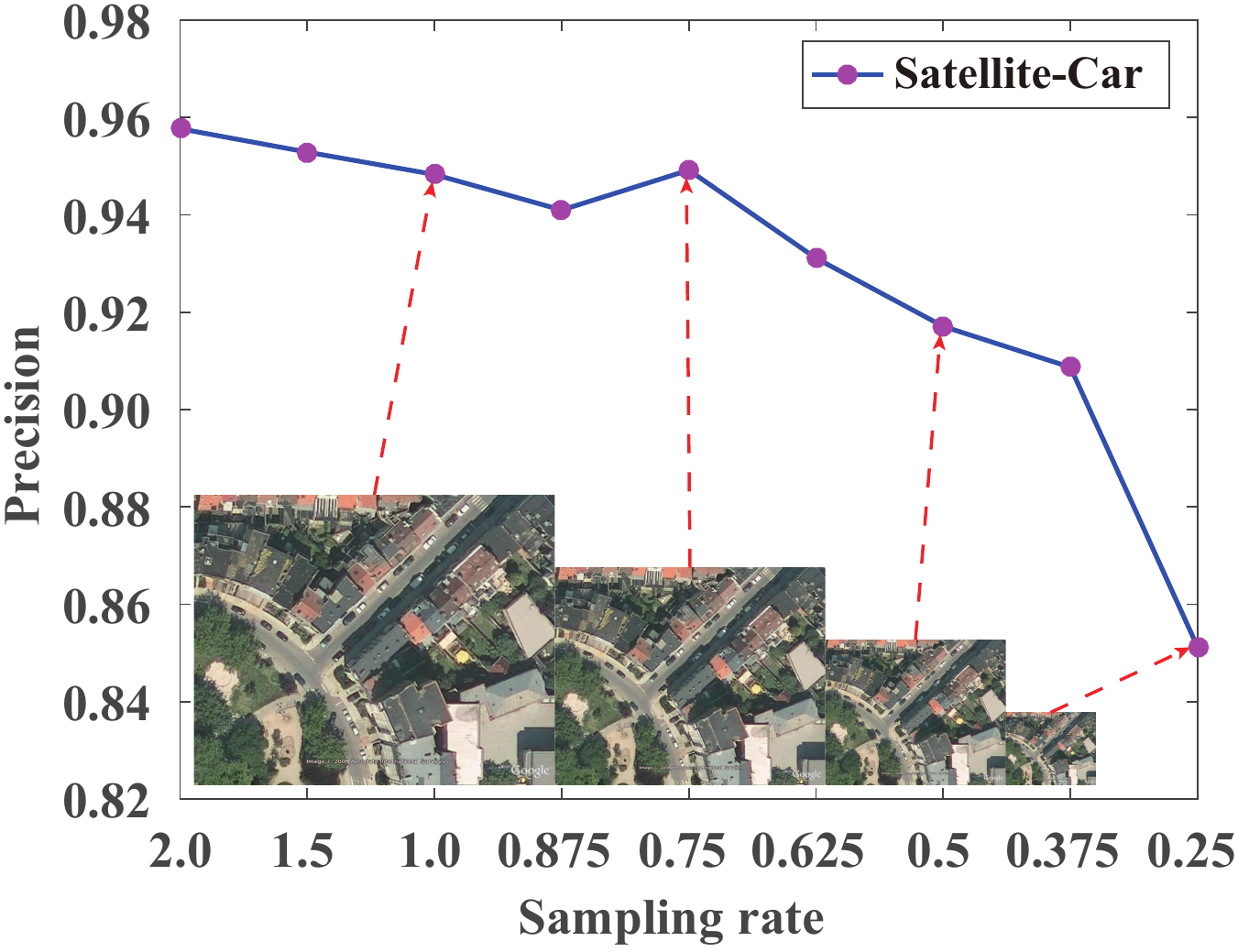}}
\subfigure[Airplane dataset]
          {\includegraphics[width=0.24\textwidth]{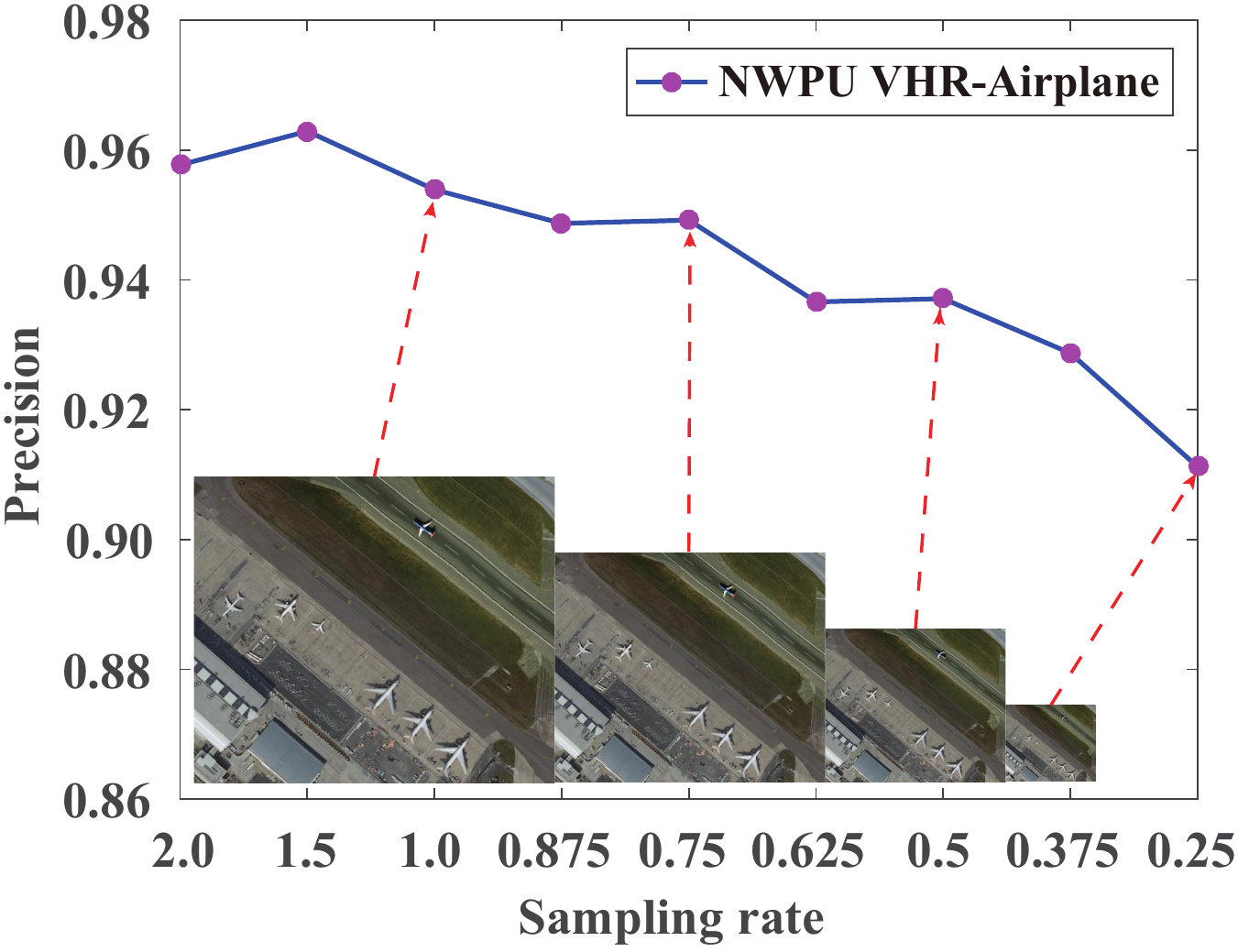}}
\caption{Sensitivity investigation to different spatial resolution of the image.}
\label{tab:resolution}
\end{figure}

We also investigate the effects of the radial profiles (the size of convolution kernel) and the Fourier orders ($k \in \left\{ {0 \sim m} \right\}$) as well as the size of sampling windows. As observed from Figs. \ref{tab:carparam} and \ref{tab:airparam} (d-e), they are relatively insensitive in a proper range, and as a result, we select them as $m=4$ for both dataset, and $r=6$, $32\times 28$ for the satellite dataset ($r=8$, $80\times 80$ for the NWPU VHR-airplane dataset). Following the same strategy with traditional detection framework, pre-smoothing and post-smoothing are usually carried out before and after running detection algorithms, in order to make the feature locally and globally smooth. The different filter $radii \in \left\{ {0,1,2,3} \right\}$ are selected for smoothing and the experimental results are given in Figs. \ref{tab:carparam} and  \ref{tab:airparam} (g-h). We simply set the \textit{radius} for both pre-smoothing and post-smoothing as 1, as they are relatively insensitive to the different radius. In the test phase, the pyramid factor plays an important role, as displayed in Figs. \ref{tab:carparam} and  \ref{tab:airparam} (i). The eight scales per octave shows a best result, which is basically consistent with \cite{PAMI2014}. Significantly, the final detection precision would increase with the number of the weak classifier, but so does the computational cost. As a trade-off, the value is set as 2048 in our case.
\subsubsection{Towards Spatial Resolution}
The image resolution is another important factor that could degrade the detection performance, and therefore we emphatically evaluate the effects of different resolution to find a proper boundary condition for the use of the proposed ORSIm detector. In detail, we adopt the different sampling rates on the two datasets to investigate the sensitivity of detection precision. As can be seen from Fig. \ref{tab:resolution}, the performance may begin to degenerate with around 0.5 sampling rate and gradually decrease after that. It should be noted that the feature pyramid is usually an indispensable step in test phase. Therefore, these detection approaches are, in fact, not so sensitive to different spatial resolution, although the lower resolution inevitably suffers from information loss. Furthermore, Fig. \ref{fig:visual result} shows a visual example to clarify that the different scaled objects can be basically detected, demonstrating the effectiveness of the ORSIm detector to the multi-resolution images. That is not to say, however, that the proposed detector is capable of handling various variations. For that, we highlight a scene to give some false cases, as shown in Fig. \ref{fig:NMS} where the detector confuses the real airplanes and its tails with a small shadow, leading to some extra false alarms marked in red. This is actually a comparatively common phenomenon in object detection rather than due to the model's sensitive to spatial resolution of an input image \cite{zhao2017effective}. A feasible solution for this issue is to use a two-step NMS, as illustrated in Fig. \ref{fig:NMS}.
\section{Conclusion}
Object rotation is a common but challenging issue for object detection and recognition in optical remote sensing. To this end, we propose a more complete object detection framework in \textbf{o}ptical \textbf{r}emote \textbf{s}ensing \textbf{im}agery, called ORSIm detector, by introducing the discriminative rotation-invariant channel features (spatial-frequency channel features), learning-based feature refining and fast feature channel scaling technique as well as boosting-based classifier learning. Extensive experimental results indicate ORSIm detector performs better and is more robust to various deformations, compared to previous state-of-arts methods. In the future work, we will focus on tiny object detection and extend the proposed framework to an end-to-end learning framework (e.g. deep learning). Additionally, we will expand the binary classification to multi-target detection.
\section*{Acknowledgment}
The authors would like to thank the Key Laboratory of Information Fusion Technology, Ministry of Education at the University of Northwestern Polytechnical and to thank the Electrical Engineering Department at Stanford University for providing the NWPU VHR-airplane dataset and the Satellite dataset. The authors would like to express their appreciation to Prof. Piotr Doll\'ar and Dr. K. liu for providing MATLAB codes for fast pyramid feature and FourierHOG algorithms.
\bibliographystyle{IEEEbib}
\bibliography{reference}

\begin{IEEEbiography}[{\includegraphics[width=1in,height=1.25in,clip,keepaspectratio]{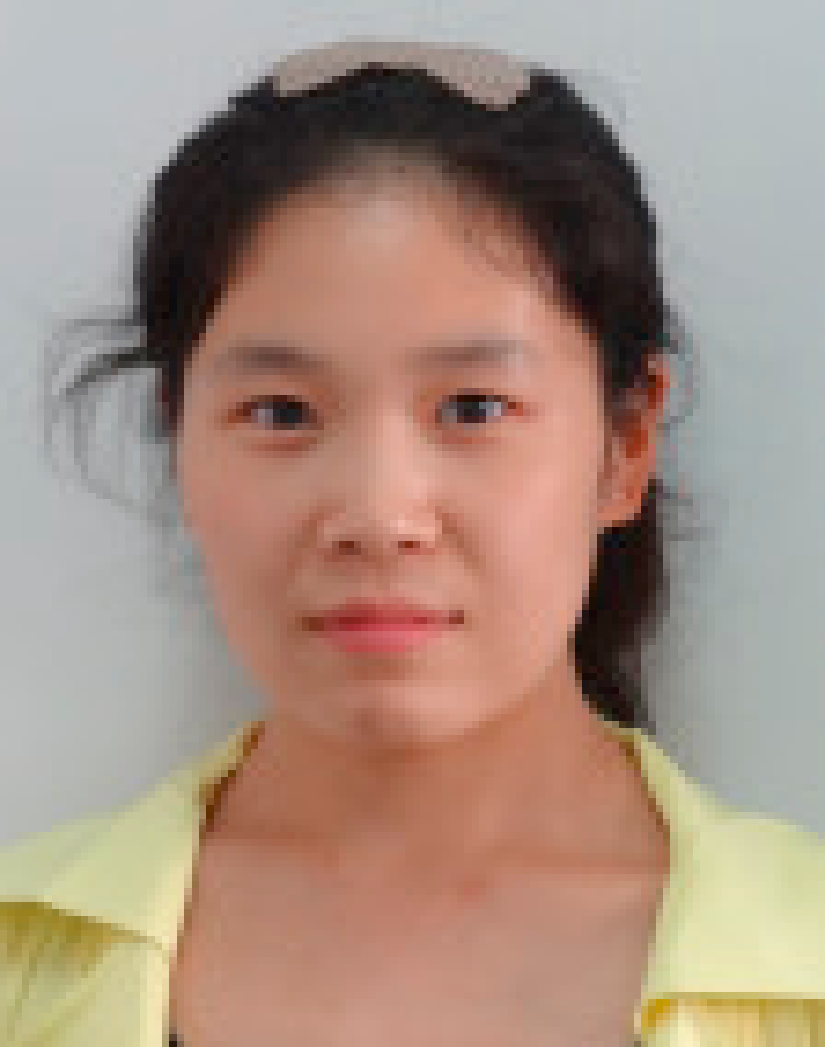}}]{Xin Wu}
(S'19) received the B.S. degree in Science and Technology of Electronic Information department from Mudanjiang Normal University, China, in 2011. She received the M.Sc. degree in Computer Science and Technology, Qingdao University, China, in 2014. She is currently pursuing her Ph.D. degree with Information and Communication Engineering in Beijing Institute of Technology (BIT), Beijing, China, since September 2014.

In 2018, she was a visiting student at the Photogrammetry and Image Analysis department of the Remote Sensing Technology Institute (IMF), German Aerospace Center (DLR), Oberpfaffenhofen, Germany, under the supervision of Dr. Jiaojiao Tian and Prof. Peter Reinartz.

Her research interests include signal / image processing, fractional fourier transform, deep learning and their applications in biometrics and geospatial object detection.
\end{IEEEbiography}

\begin{IEEEbiography}[{\includegraphics[width=1in,height=1.25in,clip,keepaspectratio]{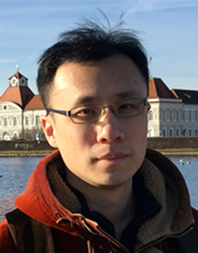}}]{Danfeng Hong}
(S'16) received the B.Sc. degree in Computer Science and Technology from the
Neusoft College of Information, Northeastern University, China, in 2012, the M.Sc. degree in Computer Vision, Qingdao University, China, in 2015. He is currently pursuing his Ph.D. degree in Signal Processing in Earth Observation, Technical University of Munich (TUM), Munich, Germany.

Since 2015, he is also a Research Associate at Remote Sensing Technology Institute (IMF), German Aerospace Center (DLR), Oberpfaffenhofen, Germany. In 2018, he was a visiting student in GIPSA-lab, Grenoble INP, CNRS, Univ. Grenoble Alpes, Grenoble, France, under the supervision of Prof. Jocelyn Chanussot.

His research interests include signal / image processing and analysis, pattern recognition, machine / deep learning and their applications in Earth Vision.
\end{IEEEbiography}
\vskip -2\baselineskip plus -1fil
\begin{IEEEbiography}[{\includegraphics[width=1in,height=1.25in,clip,keepaspectratio]{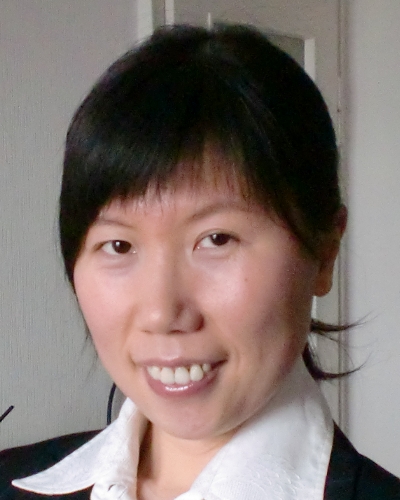}}]{Jiaojiao Tian}
received her B.S in Geo-Information Systems at the China University of Geosciences (Beijing) in 2006, M. Eng in Cartography and Geo-information at the Chinese Academy of Surveying and Mapping (CASM) in 2009, and Ph.D. degree in mathematics and computer science from Osnabrueck University, Germany in 2013.

Since September 2009, she has been working at the Photogrammetry and Image Analysis department of the Remote Sensing Technology Institute (IMF), German Aerospace Center (DLR), Oberpfaffenhofen, Germany. She is currently the Head of the 3D Modeling team. In autumn 2011, she was a guest scientist at the Institute of Photogrammetry and Remote Sensing, ETH Zurich, Switzerland. Her research interests include 3D change detection, DSM generation and quality assessment, object extraction, DSM assisted building reconstruction and classification.
\end{IEEEbiography}
\vskip -2\baselineskip plus -1fil
\begin{IEEEbiography}[{\includegraphics[width=1in,height=1.25in,clip,keepaspectratio]{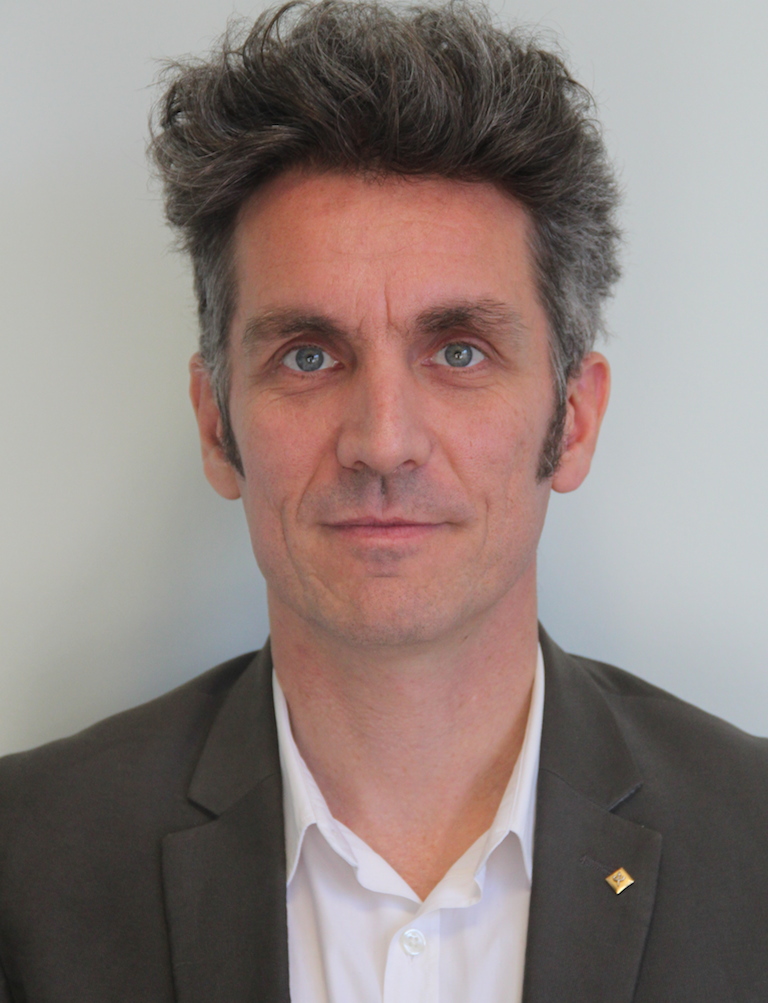}}]{Jocelyn Chanussot}
(M'04-–SM'04–-F'12) received the M.Sc. degree in electrical engineering from the Grenoble Institute of Technology (Grenoble INP), Grenoble, France, in 1995, and the Ph.D. degree from the Université de Savoie, Annecy, France, in 1998. In 1999, he was with the Geography Imagery Perception Laboratory for the Delegation Generale de l'Armement (DGA - French National Defense Department). Since 1999, he has been with Grenoble INP, where he is currently a Professor of signal and image processing. He is conducting his research at GIPSA-Lab. His research interests include image analysis, multicomponent image processing, nonlinear filtering, and data fusion in remote sensing. He has been a visiting scholar at Stanford University (USA), KTH (Sweden) and NUS (Singapore). Since 2013, he is an Adjunct Professor of the University of Iceland. In 2015-2017, he was a visiting professor at the University of California, Los Angeles (UCLA).

Dr. Chanussot is the founding President of IEEE Geoscience and Remote Sensing French chapter (2007-2010) which received the 2010 IEEE GRSS Chapter Excellence Award. He was the co-recipient of the NORSIG 2006 Best Student Paper Award, the IEEE GRSS 2011 and 2015 Symposium Best Paper Award, the IEEE GRSS 2012 Transactions Prize Paper Award and the IEEE GRSS 2013 Highest Impact Paper Award. He was a member of the IEEE Geoscience and Remote Sensing Society AdCom (2009-2010), in charge of membership development. He was the General Chair of the first IEEE GRSS Workshop on Hyperspectral Image and Signal Processing, Evolution in Remote sensing (WHISPERS). He was the Chair (2009-2011) and Co-chair of the GRS Data Fusion Technical Committee (2005-2008). He was a member of the Machine Learning for Signal Processing Technical Committee of the IEEE Signal Processing Society (2006-2008) and the Program Chair of the IEEE International Workshop on Machine Learning for Signal Processing (2009). He was an Associate Editor for the IEEE Geoscience and Remote Sensing Letters (2005-2007) and for Pattern Recognition (2006-2008). He was the Editor-in-Chief of the IEEE Journal of Selected Topics in Applied Earth Observations and Remote Sensing (2011-2015). Since 2007, he is an Associate Editor for the IEEE Transactions on Geoscience and Remote Sensing, and since 2018, he is also an Associate Editor for the IEEE Transactions on Image Processing. In 2013, he was a Guest Editor for the Proceedings of the IEEE and in 2014 a Guest Editor for the IEEE Signal Processing Magazine. He is a Fellow of the IEEE and a member of the Institut Universitaire de France (2012-2017).
\end{IEEEbiography}

\begin{IEEEbiography}[{\includegraphics[width=1in,height=1.25in,clip,keepaspectratio]{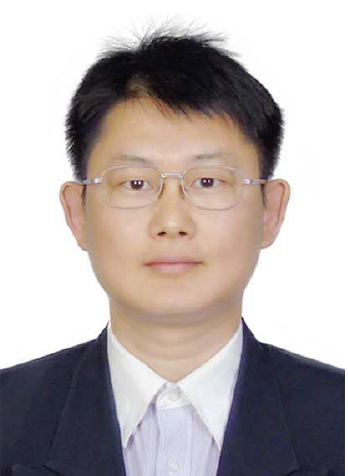}}]
{Wei Li}(S'11--M'13--SM'16) received the B.E.\ degree in telecommunications engineering from Xidian University, Xi'an, China, in 2007, the M.S.\ degree in information science and technology from Sun Yat-Sen University, Guangzhou, China, in 2009, and the Ph.D.\ degree in electrical and computer engineering from Mississippi State University, Starkville, MS, USA, in 2012. Subsequently, he spent 1 year as a Postdoctoral Researcher at the University of California, Davis, CA, USA. He is currently a Professor and Vice Dean with the College of Information Science and Technology at Beijing University of Chemical Technology, Beijing, China. His research interests include hyperspectral image analysis, pattern recognition, and data compression. Dr. Li is an active reviewer for the IEEE Transactions on Geoscience and Remote Sensing (TGRS), the IEEE Geoscience Remote Sensing Letters (GRSL), and the IEEE Journal of Selected Topics in Applied Earth Observations and Remote.
\end{IEEEbiography}
\vskip -2\baselineskip plus -1fil
\begin{IEEEbiography}[{\includegraphics[width=1in,height=1.25in,clip,keepaspectratio]{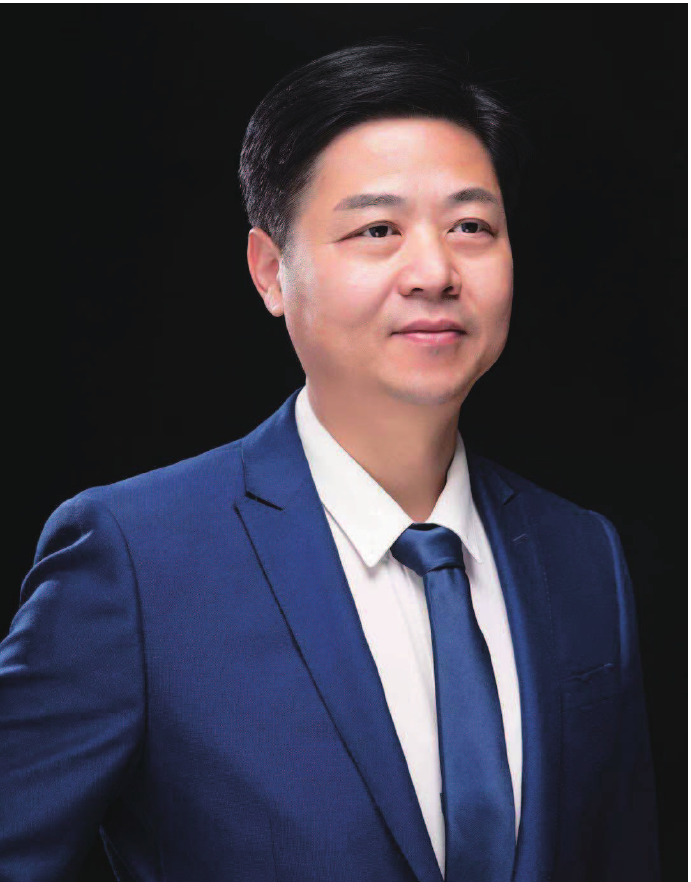}}]
{Ran Tao}(M'00–-SM'04) received the B.S. degree from the Electronic Engineering Institute of PLA, Hefei, China, in 1985, and the M.S. and Ph.D. degrees from the Harbin Institute of Technology, Harbin, China, in 1990 and 1993, respectively. He has been a Senior Visiting Scholar with the University of Michigan, Ann Arbor, MI, USA, and the University of Delaware, Newark, DE, USA, in 2001 and 2016, respectively. He is currently a Professor with the School of Information and Electronics, Beijing Institute of Technology, Beijing, China.

His current research interests include fractional Fourier transform and its applications, theory, and technology for radar and communication systems. He has 3 books and more than 100 peer-reviewed journal articles. He is a Fellow of the Institute of Engineering and Technology and the Chinese Institute of Electronics. He was a recipient of the National Science Foundation of China for Distinguished Young Scholars in 2006, and a Distinguished Professor of Changjiang Scholars Program in 2009. He has been a Chief-Professor of the Creative Research Groups with the National Natural Science Foundation of China since 2014, and he was a Chief-Professor of the Program for Changjiang Scholars and Innovative Research Team in University during 2010–-2012. He is currently the Vice-Chair of the IEEE China Council. He is also the Vice-Chair of the International Union of Radio Science (URSI) China Council and a member of Wireless Communication and Signal Processing Commission, URSI. He was a recipient of the First Prize of Science and Technology Progress in 2006 and 2007, and the First Prize of Natural Science in 2013, both awarded by the Ministry of Education.
\end{IEEEbiography}
\end{document}